%% file: main.tex
\documentclass[11pt, letterpaper, logo, onecolumn, copyright]{minimax}
\usepackage[authoryear, sort&compress, round]{natbib}
\usepackage{subcaption}
\usepackage{multirow}
\usepackage{array}
\usepackage{float}
\usepackage{adjustbox}
\usepackage{algorithm}
\usepackage{algorithmic}
\usepackage{cleveref}
\usepackage{placeins}
\input{math_commands.tex}

\theoremstyle{plain}

\theoremstyle{definition}

\theoremstyle{remark}

\newcolumntype{L}[1]{>{\raggedright\let\newline\\\arraybackslash\hspace{0pt}}m{#1}}
\newcolumntype{C}[1]{>{\centering}m{#1}}
\newcolumntype{R}[1]{>{\raggedleft\let\newline\\\arraybackslash\hspace{0pt}}m{#1}}

\newcommand{\method}{\textsc{MSA}}
\newcommand{\sg}{\mathrm{stopgrad}}

\renewcommand{\algorithmiccomment}[1]{\hfill\textcolor{gray}{// #1}}

\title{MiniMax Sparse Attention}

\reportnumber{}
\paperurl{}
\correspondingauthor{}
\author[1,2]{Xunhao Lai}
\author[1]{Weiqi Xu}
\author[1]{Yufeng Yang}
\author[3]{Qiaorui Chen}
\author[1,4]{Yang Xu}
\author[1,5]{Lunbin Zeng}
\author[1,4]{Xiaolong Li}
\author[1]{Haohai Sun}
\author[1]{Haichao Zhu}
\author[1,2]{Vito Zhang}
\author[1]{Jinkai Hu}
\author[1]{Jiayao Li}
\author[1,6]{Rui Gao}
\author[1]{Zekun Li}
\author[1]{Songquan Zhu}
\author[1,7]{Jingkai Zhou}
\author[1]{Pengyu Zhao}

\affil[1]{MiniMax}
\affil[2]{Peking University}
\affil[3]{NVIDIA}
\affil[4]{Zhejiang University}
\affil[5]{Huazhong University of Science and Technology}
\affil[6]{Nanjing University}
\affil[7]{Hangzhou Dianzi University}

\begin{abstract}
\input{sections/abstract}

\end{abstract}

\vspace{\baselineskip}
\vspace{\baselineskip}
\vspace{\baselineskip}

\begin{document}

\maketitle
\begin{figure}[!ht]
\centering
\includegraphics[width=0.95\linewidth]{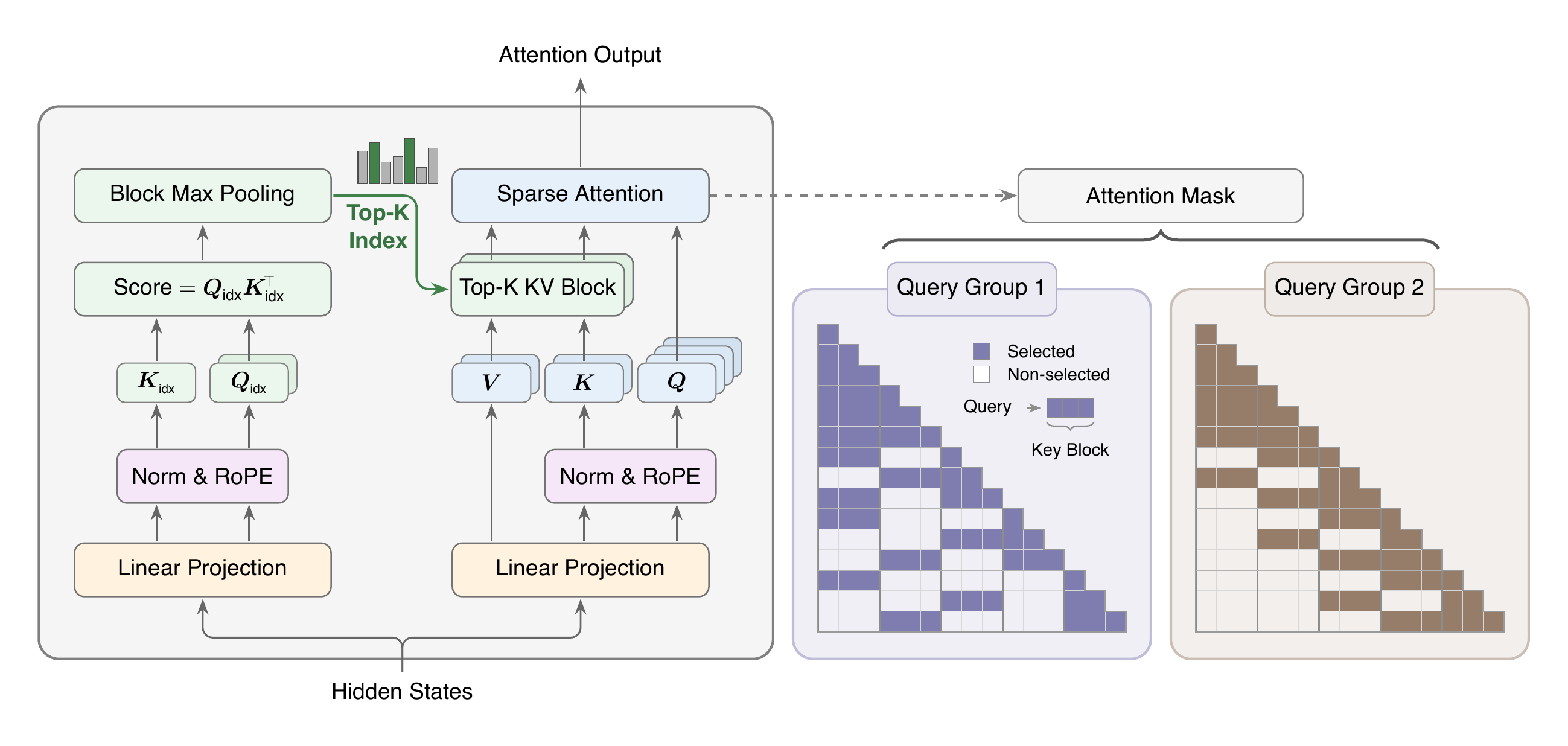}
\caption{Overview of \method{}. The Index Branch (left) scores the full causal context with a single lightweight head and selects, for each query and GQA group, a set $\gI$ of $k$ key blocks; the local block is always included regardless of its score. The Main Branch (right) attends only to the selected blocks and produces the layer output. During training, a KL loss aligns the index distribution with the group-averaged Main Branch distribution on the selected blocks, and the Index Branch gradient is detached from the Main Branch.}
\label{fig:msa-arch}
\end{figure}
\clearpage

\input{sections/introduction}
\input{sections/preliminary}
\input{sections/method}
\input{sections/infrastructure}
\input{sections/experiment}
\input{sections/related_works}
\input{sections/conclusion}

\bibliography{references}

\newpage
\input{sections/appendix}

\end{document}

%% file: math_commands.tex
\usepackage{amsmath,amsfonts,bm}

\def\Figref#1{Figure~\ref{#1}}

\def\Secref#1{Section~\ref{#1}}

\def\eqref#1{equation~\ref{#1}}
\def\Eqref#1{Equation~\ref{#1}}

\def\Algref#1{Algorithm~\ref{#1}}

\def\Tabref#1{Table~\ref{#1}}

\def\1{\bm{1}}

\def\vk{{\bm{k}}}

\def\vo{{\bm{o}}}

\def\vq{{\bm{q}}}

\def\vv{{\bm{v}}}

\def\mK{{\bm{K}}}

\def\mO{{\bm{O}}}

\def\mQ{{\bm{Q}}}

\def\mS{{\bm{S}}}

\def\mV{{\bm{V}}}
\def\mW{{\bm{W}}}
\def\mX{{\bm{X}}}

\DeclareMathAlphabet{\mathsfit}{\encodingdefault}{\sfdefault}{m}{sl}
\SetMathAlphabet{\mathsfit}{bold}{\encodingdefault}{\sfdefault}{bx}{n}

\def\gB{{\mathcal{B}}}

\def\gI{{\mathcal{I}}}

\newcommand{\Ls}{\mathcal{L}}
\newcommand{\R}{\mathbb{R}}

\newcommand{\softmax}{\mathrm{softmax}}

\newcommand{\KL}{D_{\mathrm{KL}}}



%% file: sections/abstract.tex
Ultra-long-context capability is becoming indispensable for frontier LLMs: agentic workflows, repository-scale code reasoning, and persistent memory all require the model to jointly attend over hundreds of thousands to millions of tokens—yet the quadratic cost of softmax attention makes this untenable at deployment scale. We introduce MiniMax Sparse Attention (\method{}), a blockwise sparse attention built upon Grouped Query Attention (GQA).
A lightweight Index Branch scores key–value blocks and independently selects a Top-$k$ subset for each GQA group, enabling group-specific sparse retrieval while maintaining efficient block-level execution; the Main Branch then performs exact block-sparse attention over only the selected blocks. Designed around a principle of simplicity and scalability, \method{} is deliberately streamlined, making it straightforward to deploy efficiently across a broad range of GPUs.
To translate sparsity into practical speedups, we co-design \method{} with a GPU execution path that uses exp-free Top-$k$ selection and KV-outer sparse attention to improve tensor-core utilization under block-granular access. On a 109B-parameter model with native multimodal training, \method{} performs on par with GQA while reducing per-token attention compute by $28.4\times$ at 1M context. Paired with our co-designed kernel, \method{} achieves $14.2\times$ prefill and $7.6\times$ decoding wall-clock speedups on H800. Our inference kernel is available at: \url{https://github.com/MiniMax-AI/MSA}. A production-grade natively multimodal model powered by \method{} has been publicly released at: \url{https://huggingface.co/MiniMaxAI/MiniMax-M3}.

%% file: sections/introduction.tex
\section{Introduction}
\label{sec:introduction}

Large language models (LLMs) are rapidly shifting from short, single-turn interactions to long-horizon agentic workflows that span hundreds of interleaved reasoning and action steps—writing and deploying production code, navigating the open web, orchestrating diverse tools, and producing structured documents~\citep{openai2025gpt5,anthropic2025claude46,google2025gemini31,deepseekai2026v4,moonshot2026kimi26,zhipu2026glm51}. However, the ultra-long contexts these tasks demand impose severe compute and memory bottlenecks on both training and inference, with quadratic-cost softmax attention being the primary culprit, further amplified by the latency and throughput constraints of production-scale deployment.

Context length is a critical scaling dimension for LLMs, where trading off model quality against efficiency remains a formidable challenge. The community is actively pushing the Pareto frontier on this front. Hybrid architectures~\citep{minimaxm1,qwen35blog} replace a subset of softmax attention layers with efficient alternatives such as linear attention~\citep{kimiteam2025kimilinearexpressiveefficient,yang2025gateddeltanetworksimproving,gu2023mamba} or sliding window attention~\citep{openai2025gptoss120bgptoss20bmodel,coreteam2026mimov2flashtechnicalreport}.  Alternatively, another line of work attempts to sparsify softmax attention~\citep{deepseekai2025deepseekv32pushingfrontieropen,deepseekai2026v4,minicpmteam2025minicpm4ultraefficientllmsend,lu2025mobamixtureblockattention} itself to break the computational bottleneck.

We introduce MiniMax Sparse Attention (MSA), designed following Occam's razor: after extensive ablation, we retain only the essential components. MSA follows the sparse softmax attention paradigm to maximally reuse existing software and hardware infrastructure. 
We adopt blockwise token selection with a smaller top-$k$, enabling efficient execution across a wider range of GPU architectures while relaxing the head-dimension constraints imposed by prior designs. Concretely, an ultra-lightweight Index Branch selects, for each attention group, the top-$k$ blocks via max-pooling scoring, while always retaining the most recent block to ensure training stability. 

Turning MSA's theoretical sparsity into practical end-to-end speedups requires co-designing the algorithm with its GPU execution path. To this end, we design an exp-free TopK kernel specialized for the small-k regime, leveraging the blockwise indexer to bypass unnecessary softmax computation before selection. For the main attention branch, we organize sparse attention in a KV-outer order: selected KV blocks gather their associated queries and concatenate them to fill tensor-core MMAs, using pre-scheduled chunking with a two-phase combine to handle highly skewed block popularity without atomic updates. For training, we further fuse the auxiliary LSE computation required by the sparse KL loss into the forward pass and employ persistent load balancing in the backward pass. 

To validate that MSA preserves both textual and multimodal capabilities, we compare it against Grouped Query Attention on a 109B-parameter Mixture of Experts (MoE) model trained from scratch with a 3T-token budget. MSA matches GQA on downstream benchmarks while delivering $14.2\times$ prefill and $7.6\times$ decoding speedups at 1M context length.

\noindent \textbf{Main contributions.} 
\begin{itemize}[itemsep=6pt]
    \item We propose MSA, a minimal, scalable, and accelerated blockwise sparse attention mechanism that supports both training from scratch and near-lossless conversion from pretrained GQA checkpoints.

    \item We co-design efficient training and inference kernels that turn MSA's theoretical compute savings into real wall-clock speedups at scale.

    \item We perform extensive ablations scaling up to a 109B-parameter MoE model with native multimodal training, dissecting MSA's behavior across scales and modalities.
\end{itemize}

%% file: sections/preliminary.tex
\section{Preliminary}
\label{sec:preliminary}

\subsection{Causal Attention and GQA}
\label{sec:preliminary-attention-gqa}

We write $N$ for the sequence length, $d_{\rm model}$ for the hidden dimension, and $d_h$ for the head dimension. For each query position $t$ and head $h$, causal Softmax Attention computes
\begin{equation}
\label{eq:softmax-attention}
    \vo_t^{(h)} \;=\; \sum_{i \le t} \alpha_{t,i}^{(h)} \, \vv_i^{(h)},
    \qquad
    \alpha_{t,i}^{(h)} \;=\; \frac{\exp\!\big(\langle \vq_t^{(h)}, \vk_i^{(h)} \rangle / \sqrt{d_h}\big)}{\sum_{j \le t} \exp\!\big(\langle \vq_t^{(h)}, \vk_j^{(h)} \rangle / \sqrt{d_h}\big)}.
\end{equation}
The cost of \cref{eq:softmax-attention} is $\Theta(2H_q N^2 d_h)$ FLOPs, which grows quadratically with the sequence length $N$. Grouped-Query Attention~\citep{ainslie2023gqa} uses $H_q$ query heads and reduces the number of key-value heads to $H_{kv}$, tying $G=H_q/H_{kv}$ adjacent query heads to a single shared key-value head. Thus, each key-value head defines one GQA group.

\subsection{Sparse Attention as a Two-Stage Process}
\label{sec:preliminary-sparse}

A sparse attention layer factors causal attention into an indexer that selects which keys to attend to and a sparse attention computation over the selected keys. For each query position $i$,
\begin{equation}
\label{eq:two-stage}
    \gI_i \;=\; \mathrm{Index}_\phi\!\big(\vq_i, \mK_{\le i}\big),
    \qquad
    \vo_i \;=\; \mathrm{Attn}\!\big(\vq_i, \mK[\gI_i], \mV[\gI_i]\big),
\end{equation}
where $\mathrm{Index}_\phi$ is parameterized by $\phi$ (empty for fixed-rule indexers; learned for trainable ones), $\gI_i \subseteq \{1, \dots, i\}$ denotes the selected index set, and $\mathrm{Attn}$ denotes standard scaled dot-product softmax attention restricted to this index set. We call the first stage the \emph{Index Branch} and the second the \emph{Main Branch}. In multi-head attention, each query, specified by a position $i$ and a query head $h$, can select a different key/value index set, written as $\gI_i^{(h)}$; \cref{eq:two-stage} omits the head index only for notational simplicity.

\subsection{GQA-Based Block Sparse Attention}
\label{sec:preliminary-gqa-block-sparse}

Per-head token-level selection offers the finest granularity, but such fine-grained computation is difficult to map efficiently to GPU matrix operations. For efficiency, sparse attention built on GQA can share the index result within each GQA group. Let $\mathcal{H}_r$ denote the $G$ query heads served by the $r$-th key-value head. The group-shared index set can be written as
\begin{equation}
\label{eq:prelim-group-shared-support}
    \gI_i^{(r)} = \gI_i^{(h)} = \gI_i^{(h')},
    \qquad
    h,h' \in \mathcal{H}_r .
\end{equation}

Selecting key/value blocks rather than individual tokens reduces routing overhead and makes sparse attention more regular. For block size $B_k$, define
\begin{equation}
\label{eq:prelim-key-blocks}
    \gB_b
    =
    \{(b{-}1)B_k+1,\dots,\min(bB_k,N)\},
    \qquad
    b=1,\dots,B,\quad B=\lceil N/B_k\rceil .
\end{equation}
For query position $i$ and GQA group $r$, the set $\gI_i^{(r)} \subseteq \{1,\dots,B\}$ denotes the selected block index set.
The sparse attention output for any query head in group $r$ is then computed over the causally visible tokens in the selected blocks, using the key-value head of the same group. \method{} follows this GQA-based block sparse formulation, with the concrete indexer architecture and training objective described in the next section.

%% file: sections/method.tex
\section{\method{}}
\label{sec:method}

We introduce MiniMax Sparse Attention (\method{}), a GQA-based sparse attention mechanism with two branches, as illustrated in \Figref{fig:msa-arch}. For each query token, a lightweight \emph{Index Branch} selects a small set of key blocks from the causal context, and the \emph{Main Branch} computes softmax attention over the tokens in those blocks. The Index Branch adds only two projection matrices to standard GQA, operates at block granularity, and makes selections independently for each GQA group. We describe the architecture in \Secref{sec:method-architecture} and the training procedure in \Secref{sec:method-training}.

\subsection{Architecture}
\label{sec:method-architecture}

\method{} instantiates the two-stage sparse-attention formulation in \Secref{sec:preliminary-sparse} at GQA-group and block granularity (\Figref{fig:msa-arch}). For each query token, the Index Branch selects $k$ key blocks of size $B_k$ for each GQA group, and the Main Branch attends only to tokens in the selected blocks, whose budget is at most $kB_k$.
Let $\mX \in \R^{N \times d_{\rm model}}$ be the input hidden states. Following \Secref{sec:preliminary-attention-gqa}, we write $H_q$ and $H_{kv}$ for the number of query heads and key-value heads, respectively, so each key-value head serves $G = H_q/H_{kv}$ query heads.

\paragraph{Index Branch.} The Index Branch introduces one index query head for each GQA group and a single index key head shared across groups:
\begin{equation}
\label{eq:msa-index-proj}
\mQ^{\rm idx} = \mX \mW_q^{\rm idx} \in \R^{N \times H_{kv} \times d_{\rm idx}},
\qquad
\mK^{\rm idx} = \mX \mW_k^{\rm idx} \in \R^{N \times 1 \times d_{\rm idx}}.
\end{equation}
For query token $i$ and group $r$, the Index Branch first scores visible key tokens, then aggregates these scores to the block level. Using the block partition $\gB_1,\dots,\gB_B$ defined in \Secref{sec:preliminary-gqa-block-sparse},
\begin{equation}
\label{eq:msa-index-score}
\mS^{\rm idx,(r)}_{i,j}
\;=\;
\frac{\bigl(\mQ^{\rm idx}\bigr)^{(r)}_i \,\bigl(\mK^{\rm idx}\bigr)_j^{\top}}
     {\sqrt{d_{\rm idx}}},
\qquad
M^{\rm idx,(r)}_{i,b}
\;=\;
\max_{\substack{j \in \gB_b \\ j \le i}}
\mS^{\rm idx,(r)}_{i,j}.
\end{equation}
Here $r$ indexes the GQA group, $j \le i$ enforces causality, and blocks with no visible token are assigned score $-\infty$. The Index Branch then selects the top-$k$ block indices:
\begin{equation}
\label{eq:msa-topk}
\gI_i^{(r)}
\;=\;
\mathrm{TopK}_{b \in \{1,\dots,B\}}\!\bigl(M^{\rm idx,(r)}_{i,\cdot},\, k\bigr).
\end{equation}
Here $\mathrm{TopK}(\cdot, k)$ returns the indices of the $k$ largest blocks under $M^{\rm idx,(r)}_{i,\cdot}$. We always include the local block containing position $i$, and $\gI_i^{(r)}$ is shared by all $G$ query heads in group $r$.

\paragraph{Main Branch.} Given the block index set $\gI_i^{(r)}$ selected by the Index Branch, the Main Branch attends only to the causally visible tokens in the selected blocks. For any query head $h \in \mathcal{H}_r$, it applies standard scaled dot-product attention restricted to these tokens, using the key-value head associated with GQA group $r$:
\begin{equation}
\label{eq:msa-sparse-attn}
\mO_{i}^{(h)}
\;=\;
\softmax\Biggl(
  \frac{\mQ_{i}^{(h)}\,
        \bigl(\mK^{(r)}\!\bigl[\gI_i^{(r)}\bigr]\bigr)^{\top}}
       {\sqrt{d_h}}
\Biggr)
\mV^{(r)}\!\bigl[\gI_i^{(r)}\bigr],
\end{equation}
where $\mQ_{i}^{(h)}$ denotes the query vector at position $i$ and query head $h$, while $\mK^{(r)}$ and $\mV^{(r)}$ denote the key and value matrices of the $r$-th GQA group.
The notation $\mK^{(r)}[\gI_i^{(r)}]$ and $\mV^{(r)}[\gI_i^{(r)}]$ denotes gathering the causally visible tokens from the selected blocks.
The block index set $\gI_i^{(r)}$ is shared by all query heads in $\mathcal{H}_r$, while each head keeps its own query projection. Since the selected blocks contain at most $kB_k$ causally visible tokens, the per-query attention cost is reduced from $O(N)$ to $O(kB_k)$, which is fixed as the sequence length increases.

\subsection{Training}
\label{sec:method-training}

The top-$k$ selection in \Eqref{eq:msa-topk} is non-differentiable, so the language-modeling loss cannot train the index $Q/K$ projections $\mW^{\rm idx}_q, \mW^{\rm idx}_k$ directly. We therefore train the Index Branch with a KL alignment loss and use three mechanisms to stabilise sparse training: \textbf{Gradient Detach}, \textbf{Indexer Warmup}, and a forced \textbf{Local Block}. We describe each component below.

\paragraph{KL Loss.} The KL loss gives the Index Branch a direct learning signal by matching its scores to the Main Branch on the selected tokens. Writing $\gI_{i,\mathrm{tok}}^{(r)}=(\bigcup_{b\in\gI_i^{(r)}}\gB_b)\cap\{1,\dots,i\}$ for the causally visible tokens induced by the selected block indices, for each query position $i$ and GQA group $r$, we define the Index Branch distribution $P^{\rm idx}$ and the Main Branch teacher $P$ over this token index set:
\begin{equation}
\label{eq:msa-pp}
P^{{\rm idx},(r)}_{i,j}
= \frac{\exp(S^{{\rm idx},(r)}_{i,j})}
       {\sum_{u\in\gI_{i,\mathrm{tok}}^{(r)}}\exp(S^{{\rm idx},(r)}_{i,u})},
\qquad
P^{(r)}_{i,j}
= \frac{1}{G}\sum_{\ell\in\mathcal{H}_r}
  \frac{\exp(S^{(\ell)}_{i,j})}
       {\sum_{u\in\gI_{i,\mathrm{tok}}^{(r)}}\exp(S^{(\ell)}_{i,u})},
\qquad j\in\gI_{i,\mathrm{tok}}^{(r)},
\end{equation}
where $S^{{\rm idx},(r)}_{i,j} = (\mQ^{\rm idx})^{(r)}_i(\mK^{\rm idx})_j^{\top}/\sqrt{d_{\rm idx}}$ is the token-level index score, and $S^{(\ell)}_{i,j} = \mQ^{(\ell)}_i(\mK^{(r)}_j)^{\top}/\sqrt{d_h}$ is the Main Branch score for query head $\ell\in\mathcal{H}_r$. The teacher $P$ averages the per-head Main Branch distributions at the probability level. The indexer is then trained to match $P$, averaged over all query positions and GQA groups:
\begin{equation}
\label{eq:msa-kl}
\Ls_{\rm KL}
= \frac{1}{NH_{kv}}
  \sum_{i=1}^{N}\sum_{r=1}^{H_{kv}}
  \KL\bigl(\sg(P^{(r)}_{i,\cdot}) \,\|\, P^{{\rm idx},(r)}_{i,\cdot}\bigr),
\end{equation}
where $N$ is the sequence length, and the teacher distribution $P^{(r)}_{i,\cdot}$ is detached from gradient computation. This auxiliary loss aligns the index distribution with the Main Branch attention pattern, making the subsequent block selection semantically meaningful.

\paragraph{Gradient Detach.} To isolate the auxiliary objective from the backbone, we apply stop-gradient to the Index Branch input:
\begin{equation}
\label{eq:msa-detach}
\mQ^{\rm idx} \;=\; \sg(\mX)\mW^{\rm idx}_q,
\qquad
\mK^{\rm idx} \;=\; \sg(\mX)\mW^{\rm idx}_k.
\end{equation}
The teacher $P$ in \Eqref{eq:msa-pp} is detached, so $\Ls_{\rm KL}$ leaves the Main Branch projections untouched; \Eqref{eq:msa-detach} further prevents it from reaching the backbone through $\mX$. Under this rule, $\Ls_{\rm KL}$ updates only $\mW^{\rm idx}_q$ and $\mW^{\rm idx}_k$, making the KL a clean alignment signal for the indexer.

\paragraph{Indexer Warmup.} We use a two-stage training schedule to initialise the Index Branch and avoid early random selections. During the first few iterations, the model runs full attention in both branches and trains the newly added index projections with $\Ls_{\rm KL}$. After warmup, the model switches to sparse attention, and $\Ls_{\rm KL}$ is computed over the top-$k$ selected positions. The same schedule is used when sparsifying a pretrained full-attention checkpoint, which helps align the newly added index projections before they control Main Branch routing.

\paragraph{Local Block.} For each query position $i$ and GQA group $r$, the local block containing $i$ is always selected as part of $\gI_i^{(r)}$ during both training and inference. This fixed allocation reserves one block slot and leaves the remaining slots to be chosen by the Index Branch, preventing degenerate selections that omit the query's immediate neighbourhood.

The complete layer-level training procedure is summarised in \Algref{alg:msa-train}.

\begin{algorithm}[t]
\caption{One \method{} layer: training forward and the auxiliary KL loss. The layer returns its output and per-layer $\Ls_{\rm KL}$; the model loss $\Ls = \Ls_{\rm LM} + \lambda\sum_{\rm layers}\Ls_{\rm KL}$ is assembled by the training loop.}
\label{alg:msa-train}
\begin{algorithmic}[1]
\REQUIRE hidden states $\mX \in \R^{N \times d_{\rm model}}$; block size $B_k$,
number of selected blocks $k$.
\STATE $\mQ, \mK, \mV \leftarrow \mX\mW_q,\, \mX\mW_k,\, \mX\mW_v$
       \COMMENT{$(N,H_q,d_h),(N,H_{kv},d_h),(N,H_{kv},d_h)$}
\STATE $\mQ^{\rm idx}, \mK^{\rm idx} \leftarrow
       \sg(\mX)\mW^{\rm idx}_q,\, \sg(\mX)\mW^{\rm idx}_k$
       \COMMENT{$(N,H_{kv},d_{\rm idx}),(N,1,d_{\rm idx})$; detached}
\STATE $M^{\rm idx} \leftarrow \mathrm{BlockMaxPool}(\mQ^{\rm idx}, \mK^{\rm idx}, B_k)$
       \COMMENT{$(N,H_{kv},B)$; per-group, causal}
\STATE $\gI \leftarrow \mathrm{TopK}(M^{\rm idx},\, k)$
       \COMMENT{selected block indices; local block included}
\STATE $\mO \leftarrow \mathrm{TopKAttn}(\mQ, \mK, \mV, \gI)$
       \COMMENT{$(N,H_q,d_h)$; attends to selected blocks}
\STATE $\mathrm{output} \leftarrow \mO\mW_o$
       \COMMENT{$(N,d_{\rm model})$}
\STATE $\Ls_{\rm KL} \leftarrow
       \mathrm{KLdiv}(\mQ^{\rm idx}, \mK^{\rm idx},\, \sg(\mQ), \sg(\mK),\, \gI)$
       \COMMENT{over tokens induced by $\gI$}
\STATE \textbf{return}~$\mathrm{output},\ \Ls_{\rm KL}$
\end{algorithmic}
\end{algorithm}

\subsection{Computational Complexity}
\label{sec:msa-flops}

Under the same $H_q$, $H_{kv}$, $d_h$, and sequence length $N$, the causal attention FLOPs of GQA and \method{} are
\begin{equation}
F_{\rm GQA}(N)= 2 H_q d_h N^2,
\qquad
F_{\method{}}(N)= \underbrace{H_{kv} d_{\rm idx} N^2}_{\text{Index Branch}}
 + \underbrace{4 H_q d_h Nk B_k}_{\text{Main Branch}} .
\end{equation}
GQA scales its main attention path with the full context length, whereas \method{} uses a fixed selection budget $kB_k$ plus a lightweight index computation; the FLOPs gap therefore grows with $N$ when $kB_k \ll N$ and $H_{kv}d_{\rm idx} \ll H_qd_h$.

%% file: sections/infrastructure.tex
\section{Kernel Design}
\label{sec:infrastructure}

This section describes the GPU kernels used in our sparse prefill implementation, including the index TopK kernel, the KV-outer sparse attention forward, and the sparse KL loss backward.

\subsection{Index \& TopK}

\paragraph{Exp-free selection.}
To efficiently select the top-$k$ KV blocks, the index module ranks the index scores $s$ directly. Since softmax is order-preserving, the relative ordering of scores is preserved ($s_i \le s_j \iff \mathrm{softmax}(s)_i \le \mathrm{softmax}(s)_j$), leaving the top-$k$ indices unchanged. The forward pass, therefore, bypasses the max/exp/sum steps of softmax and passes raw scores directly to selection.

\paragraph{Per-thread register top-$k$.}
The block size $B_k$ and selection size $k$ are co-designed with the top-$k$ kernel: a large $B_k$ increases attention arithmetic intensity (Section~\ref{sec:kernel:sparse_attn}), and a small $k$ at this $B_k$ keeps both the per-row candidate block count $B$ and $k$ below the sweet spot of general-purpose top-$k$ kernels, which amortize multi-pass bucketing over large $B$ (radix selection) or scale as $O(B \log^2 B)$ (bitonic sort). We adopt $B_k = 128$, $k = 16$. Each of the warp's 32 lanes streams a 1/32 stride of the input row and maintains a $k$-element min-heap in shared memory. The heap root is cached in a register, and insertions are performed with deferred writes. Finally, a $k$-round shuffle merge combines the 32 local TopK results. The shared-memory layout maps each lane to a fixed bank, avoiding conflicts.

\paragraph{Benchmark.}
We compare against \texttt{torch.topk} and the TileLang~\citep{wang2025tilelangcomposabletiledprogramming} radix-select top-$k$ on an H800 GPU with \texttt{fp32} inputs and unsorted outputs; latencies are the median of $50$ post-warmup iterations. Table~\ref{tab:topk_latency} shows that our specialized kernel is fastest in all tested settings, with the largest gains at the deployed setting $k = 16$.

\begin{table}[h]
\centering
\begin{tabular}{rrrrrrrr}
\toprule
Seq. Len. $N$ & Blocks $B$ & $k$ & \texttt{torch} & TileLang & Ours & vs.\ \texttt{torch} & vs.\ TileLang \\
\midrule
$128$K & $1024$ & $16$ & $3970$  & $2864$  & $779$   & $5.1\times$ & $3.7\times$ \\
$128$K & $2048$ & $32$ & $5378$  & $3630$  & $1991$  & $2.7\times$ & $1.8\times$ \\
$512$K & $4096$ & $16$ & $33810$ & $17779$ & $7880$  & $4.3\times$ & $2.3\times$ \\
$512$K & $8192$ & $32$ & $57659$ & $26100$ & $21326$ & $2.7\times$ & $1.2\times$ \\
\bottomrule
\end{tabular}
\caption{Top-$k$ latency ($\mu$s) for \texttt{fp32} inputs of shape $(N, B)$, with rows processed independently. The deployed setting uses $B_k = 128$, $k = 16$, while for reference we also report $k = 32$ with $B_k = 64$. All implementations produce identical index sets.}

\label{tab:topk_latency}
\end{table}
\FloatBarrier

\subsection{Sparse Attention}
\label{sec:kernel:sparse_attn}

We revisit the choice of iteration order under sparse prefill with equal query and key/value lengths. Let $H_q$, $H_{kv}$, $G = H_q / H_{kv}$, $d_h$, $N$, $B_k$, and $k$ denote the number of query heads, key-value heads, GQA ratio, head dimension, sequence length, KV block size, and number of blocks selected per query.  For simplicity, the IO estimates below assume 2-byte elements (bfloat16-sized traffic). Our kernels also support fp8; using fp8 rescales the absolute IO volume but leaves the comparison between Q-outer and KV-outer iteration unchanged.

Iterating queries on the outer loop gives
\begin{align}
\mathrm{FLOPs} &= 4\, H_q\, N\, d_h\, k\, B_k, \\
\mathrm{IO}    &= \underbrace{2 \cdot 2 \cdot H_q\, N\, d_h}_{\text{read}(\mQ)+\text{write}(\mO)}
                + \underbrace{2 \cdot 2 \cdot H_{kv}\, N\, k\, B_k\, d_h}_{\text{read}(\mK+\mV)},
\end{align}
hence $\mathrm{FLOPs}/\mathrm{IO} \approx G$.

Iterating KV blocks on the outer loop and gathering the queries that selected each block requires an intermediate output buffer:
\begin{align}
\mathrm{FLOPs} &= 4\, H_q\, N\, d_h\, k\, B_k, \\
\mathrm{IO}    &= \underbrace{2 \cdot 2 \cdot H_{kv}\, N\, d_h}_{\text{read}(\mK+\mV)}
                + \underbrace{2 \cdot 2 \cdot H_q\, N\, k\, d_h}_{\text{read}(\mQ)+\text{write}(\mO_\text{buf})}
                + \underbrace{2 \cdot H_q\, N\, (k{+}1)\, d_h}_{\text{read}(\mO_\text{buf})+\text{write}(\mO)},
\end{align}
hence $\mathrm{FLOPs}/\mathrm{IO} \approx \tfrac{2}{3} B_k$.

Since $\tfrac{2}{3} B_k \gg G$ in practice, we choose KV-outer iteration with Q gather to maximize arithmetic intensity. The kernel executes as a persistent grid over $(\textit{kv\_block}, \textit{kv\_head})$ tiles. For each tile, a reverse sparse index from the TopK selection identifies the relevant query positions. These queries are loaded into shared memory via TMA copies, one per query token, dispatched in parallel by the 32 lanes of a warp.

\paragraph{Pre-scheduled tile chunking.}
A direct one-CTA-per-tile mapping is dominated by sink rows---a single early KV block selected by nearly every query---and the same hotspot pattern can arise on any popular KV block. A GPU scheduler kernel therefore splits each KV tile along its query dimension into chunks of at most $\sim\!2 k B_k$ queries each, fanning hot tiles across many CTAs that share the same $\mathbf{K}/\mathbf{V}$ load.
Because each query's $k$ partials are now produced by $k$ CTAs, the scheduler also preassigns each (query, chunk) pair a slot $s \in [0, k)$ in $\mathbf{O}_\text{buf}$---packed with the query index $i$ into a $32$-bit handle---so the attention kernel writes its partial to the preassigned offset without atomics. The combine kernel reads a per-query slot count to know how many partials to merge.

\paragraph{Two-phase forward.}
The KV-outer split forbids inline softmax normalization since each query's $k$ partials are produced by $k$ different CTAs. The forward is therefore split into two kernels separated by HBM buffers $\mathbf{O}_\text{buf} \in \mathbb{R}^{k \times n \times H_q \times d}$ (locally normalized partial outputs) and $\mathrm{LSE}_\text{buf} \in \mathbb{R}^{k \times n \times H_q}$ (per-partial logsumexps). The attention kernel runs the worklist above and writes each partial to its preassigned slot. The combine kernel reads the valid slots of each query, computes $a = \max_s \mathrm{LSE}_s$ and $\mathrm{LSE}[i, h] = a + \log \sum_s \exp(\mathrm{LSE}_s - a)$, then forms normalized split-K weights $w_s = \exp(\mathrm{LSE}_s - \mathrm{LSE}[i, h])$. It outputs $\mathbf{O}[i, h] = \sum_s w_s\, \mathbf{O}_\text{buf}[s, i, h]$ together with the final $\mathrm{LSE}[i, h]$. The two kernels use Programmatic Dependent Launch to hide the inter-kernel launch
latency.

\paragraph{Query concatenation.}
KV-outer iteration often associates each KV tile with only a few to a few tens of query positions. Processing these positions one at a time would under-fill the score MMA: with $G = 16$, a single query position contributes only $G$ query heads, yielding an MMA $M$ dimension of only 16. Under Q-outer iteration, query positions cannot be concatenated along the sequence dimension because they generally select different KV subsets. Under KV-outer iteration, however, all gathered positions for a given tile share the same KV operands. The kernel, therefore, packs $\lceil 128/G \rceil$ query positions together with their $G$ associated query heads, all under the same KV head, into a $128 \times 128$ score MMA.

\subsection{Sparse KL Loss}
\label{sec:kernel:sparse_kl}

\paragraph{LSE fusion.}
In our initial implementation, we utilized a dedicated kernel to compute the KL divergence forward pass, storing $\mathrm{LSE}_{\rm main}$ and $\mathrm{LSE}_{\rm idx}$ to facilitate backpropagation. However, since the KL loss only affects the backward gradient, we optimize this by emitting these LSE values directly to global memory during the main pass, allowing us to skip the KL loss forward pass entirely. Additionally, during the index branch computation, we save the per-block LSEs and perform a reduction over the top-$k$ blocks to obtain $\mathrm{LSE}_{\rm idx}$. The backward kernel then loads these scalars directly into the softmax, eliminating the redundant forward computation.

\paragraph{Dynamic load balancing.}
Per-tile work varies by orders of magnitude under variable-length sequences and data-dependent sparsity. The kernel runs as a persistent grid in which CTAs claim work through a global atomic counter; each tile is partitioned along its gathered-query dimension into sub-tiles whose count scales with the per-tile query count, subject to a minimum sub-tile granularity that amortizes per-sub-tile overhead.

%% file: sections/experiment.tex
\section{Experiment}
\label{sec:experiment}

This section reports two 109B-scale experiments used to validate the final \method{} design on a native multimodal model trained on a mixture of text and image/video data. The first trains a native \method{} model from scratch, which we denote as \textbf{\method{}-PT}. The second starts from a Full-Attention checkpoint and continues pretraining after replacing dense attention with \method{}, which we denote as \textbf{\method{}-CPT}. Both models use the same architecture family as the Full-Attention baseline, but replace dense attention with the \method{} layer.

\subsection{Setup}
\label{sec:experiment-setup}

\paragraph{Model Structure.} All models use the same 41-layer MoE backbone, with approximately 109B total parameters and 6B activated parameters per token. The first three layers are dense layers, and the remaining 38 layers are MoE layers. The model uses a 200K-token vocabulary and hidden size $d_{\rm model}=3072$. Each attention module uses MSA with 64 query heads, 4 KV heads, head dimension 128, and RoPE dimension 64. Each MoE layer uses 128 routed experts, 1 shared expert, and top-4 routed expert selection. During sparse training and evaluation, both \method{} models use block size \(B_k=128\) and keep \(k=16\) key-value blocks per query and GQA group.

\paragraph{Training Budget.} All models are trained under a total budget of 3T tokens. \method{}-PT is trained from scratch: after a 40B-token indexer warmup, it remains in sparse training for the rest of pretraining. \method{}-CPT starts from a GQA Full-Attention checkpoint trained on 2.6T tokens. We then replace dense attention with \method{} and continue training for 400B tokens: the first 40B tokens are used for indexer warmup, followed by sparse continued pretraining.

\paragraph{Evaluations.} We evaluate Full, \method{}-PT, and \method{}-CPT on the same pretraining evaluation suite using matched checkpoints under the same training budget. For general reasoning and question answering, we use MMLU~\citep{hendrycks2021measuring}, MMLU-Pro~\citep{NEURIPS2024_ad236edc}, BBH~\citep{suzgun2022challengingbigbenchtaskschainofthought}, GPQA Hard~\citep{rein2023gpqa}, ARC Challenge~\citep{clark2018thinksolvedquestionanswering}, TriviaQA~\citep{joshi-etal-2017-triviaqa}, and WinoGrande~\citep{Sakaguchi_2020}. For math and code, we use GSM8K~\citep{cobbe2021trainingverifierssolvemath}, MGSM~\citep{shi2022languagemodelsmultilingualchainofthought}, MathVista~\citep{lu2024mathvista}, OlymMATH~\citep{sun2025olymmath}, HumanEval~\citep{chen2021evaluatinglargelanguagemodels}, EvalPlus~\citep{NEURIPS2023_43e9d647}, BigCodeBench~\citep{zhuo2025bigcodebench}, and MultiPL-E MBPP~\citep{cassano2023multipl}. We also evaluate multimodal capability: image benchmarks include AI2D~\citep{kembhavi2016ai2d}, ChartQA~\citep{masry-etal-2022-chartqa}, MMMU~\citep{yue2024mmmu}, OCRBench v2~\citep{fu2025ocrbenchv2improvedbenchmark}, CharXiv~\citep{wang2024charxiv}, VisualWebBench~\citep{liu2024visualwebbench}, and CVBench~\citep{NEURIPS2024_9ee3a664}, while video benchmarks include EgoSchema~\citep{mangalam2023egoschema}, LongVideoBench~\citep{NEURIPS2024_329ad516}, MLVU~\citep{Zhou_2025_CVPR}, MMVU~\citep{zhao2025mmvu}, VideoMME~\citep{fu2024videomme}, and TemporalBench~\citep{cai2024temporalbenchbenchmarkingfinegrainedtemporal}. For long-context evaluation, we use RULER~\citep{hsieh2024ruler} and HELMET~\citep{yen2025helmet}. We additionally report perplexity on downstream agent tasks, including $\tau^2$-bench~\citep{barres2025tau2}, TheAgentCompany~\citep{xu2024theagentcompany}, Humanity's Last Exam~\citep{phan2025hle}, and SWE-bench~\citep{jimenez2024swebench}.

\subsection{Training Dynamics}
\label{sec:experiment-training-dynamics}

\begin{figure}[H]
\centering
\begin{subfigure}[t]{0.48\linewidth}
  \centering
  \includegraphics[width=0.84\linewidth]{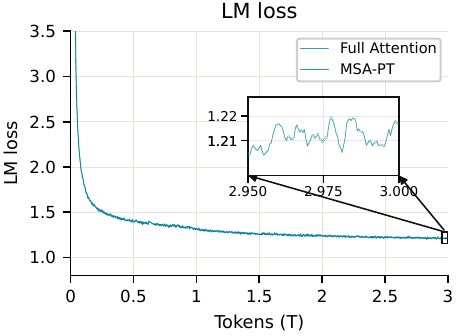}
  \caption{LM loss.}
\end{subfigure}\hfill%
\begin{subfigure}[t]{0.48\linewidth}
  \centering
  \includegraphics[width=0.84\linewidth]{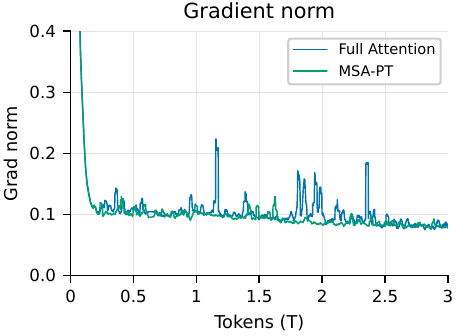}
  \caption{Gradient norm.}
\end{subfigure}
\caption{Pretraining dynamics for the experiment model. LM loss and gradient norm are shown for Full Attention and \method{}-PT over 3T training tokens. The inset in (a) zooms in on the final 50B-token window, where the two LM-loss curves remain nearly overlapping.}
\label{fig:training-dynamics-pt}
\end{figure}

\Figref{fig:training-dynamics-pt} compares native sparse pretraining with the matched full-attention run. Over the 3T-token training process, the two LM-loss curves are nearly indistinguishable, showing that \method{} does not introduce noticeable optimization degradation relative to full attention. The gradient-norm curves also remain within the same range throughout training, suggesting that \method{} does not lead to abnormal gradient fluctuations or training instability. These results indicate that training a sparse attention model is as stable as training the full-attention baseline at a large scale.

\begin{figure}[H]
\centering
\begin{subfigure}[t]{0.48\linewidth}
  \centering
  \includegraphics[width=0.84\linewidth]{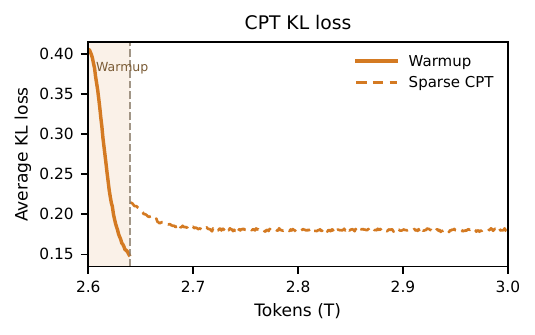}
  \caption{KL loss.}
\end{subfigure}\hfill
\begin{subfigure}[t]{0.48\linewidth}
  \centering
  \includegraphics[width=0.84\linewidth]{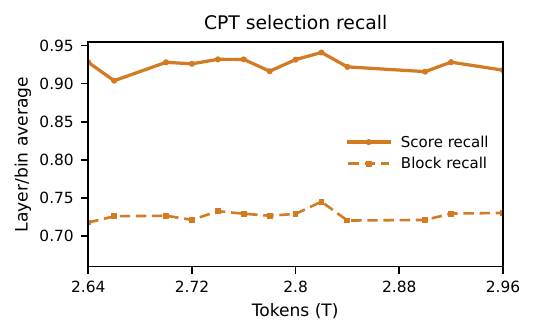}
  \caption{Selection recall.}
\end{subfigure}
\caption{Sparse continued-pretraining dynamics. (a) Average KL loss during \method{}-CPT. The solid segment denotes indexer warmup, and the dashed segment denotes sparse continued pretraining; the vertical dashed line marks the switch between the two stages. (b) Average block recall and score recall of the \method{}-CPT indexer during sparse continued pretraining.}
\label{fig:training-dynamics-cpt}
\end{figure}

\Figref{fig:training-dynamics-cpt} illustrates the transition from a trained full-attention checkpoint to sparse continued pretraining. The indexer-warmup stage rapidly reduces the KL loss before sparse attention is enabled. After switching to sparse CPT, the KL loss remains low.
For each query and GQA head, let $\gI^\star$ be the corresponding Top-$k$ block set induced by the Main Branch scores and let $\widehat{\gI}$ be the Index Branch selection. Block recall is $|\gI^\star\cap\widehat{\gI}|/|\gI^\star|$, while score recall is $\sum_{b\in\gI^\star\cap\widehat{\gI}} P_b/\sum_{b\in\gI^\star} P_b$, where $P_b$ is the Main Branch attention probability summed over tokens in block $b$. 
The block recall stays favorable, indicating reliable recovery of important blocks. The higher score recall further shows that the retrieved blocks account for most of the Main Branch attention mass.
Together, these dynamics show that warmup provides a clean conversion phase and that the CPT indexer remains well aligned during sparse continued pretraining.

\subsection{Main Results}
\label{sec:experiment-main-results}

\Tabref{tab:pretrain-representative} compares Full, \method{}-PT, and \method{}-CPT on a representative set of pretraining evaluations. Both sparse models remain broadly competitive with the Full-Attention baseline, indicating that replacing dense attention with \method{} does not substantially degrade the model's general language, reasoning, multimodal, or agent-oriented perplexity profile. The two training routes show different strengths. \method{}-PT, which learns the sparse pattern throughout pretraining, obtains the strongest results on many math, image, video, and long-context retrieval benchmarks, suggesting that native sparse pretraining can adapt the model representations to the sparse attention pattern. \method{}-CPT is more conservative: it preserves much of the Full-Attention checkpoint behavior and remains close on most text, code, and PPL evaluations, making it a practical conversion route when a trained dense checkpoint is already available. The remaining gaps are benchmark-dependent rather than concentrated in a single capability area.

\begin{table}[H]
\centering
\caption{Representative evaluation results under the 3T-token training budget. Full denotes the Full-Attention baseline, \method{}-PT denotes from-scratch sparse pretraining, and \method{}-CPT denotes sparse continued pretraining. Best per-row results are bolded; lower is better for PPL and higher is better otherwise.}
\label{tab:pretrain-representative}
{\small
\renewcommand{\arraystretch}{0.92}
\begin{adjustbox}{max width=0.95\linewidth}
\begin{tabular}{llrrr}
\toprule
Group & Benchmark & Full & \method{}-PT & \method{}-CPT \\
\midrule
\multirow{7}{*}{\textit{General}} & MMLU & 67.0 & \textbf{67.2} & 66.8 \\
 & MMLU-Pro & 38.5 & 38.8 & \textbf{39.1} \\
 & BBH & \textbf{67.7} & 66.6 & 66.1 \\
 & GPQA Hard & 25.9 & \textbf{26.3} & \textbf{26.3} \\
 & ARC Challenge & 82.7 & 82.5 & \textbf{82.9} \\
 & TriviaQA & 66.0 & 65.5 & \textbf{67.7} \\
 & WinoGrande & 58.3 & 60.9 & \textbf{62.0} \\
\midrule
\multirow{4}{*}{\textit{Math}} & GSM8K & 76.2 & \textbf{77.7} & 73.7 \\
 & MGSM & 44.1 & \textbf{46.0} & 44.2 \\
 & MathVista & 43.8 & \textbf{46.8} & 44.5 \\
 & OlymMATH Easy P@100 & 23.0 & \textbf{26.0} & 22.0 \\
\midrule
\multirow{4}{*}{\textit{Code}} & HumanEval & 61.0 & \textbf{64.0} & 57.9 \\
 & EvalPlus & 59.4 & \textbf{61.8} & 60.0 \\
 & BigCodeBench & 44.8 & 44.0 & \textbf{45.7} \\
 & MultiPL-E MBPP P@10 & \textbf{82.1} & 81.6 & 81.1 \\
\midrule
\multirow{2}{*}{\textit{Retrieval}} & RULER-8K & 79.8 & \textbf{84.2} & 77.2 \\
 & RULER-32K & 75.0 & \textbf{77.5} & 75.7 \\
\midrule
\multirow{7}{*}{\textit{Image}} & AI2D & 68.3 & \textbf{70.6} & 67.3 \\
 & ChartQA & 75.0 & \textbf{75.4} & 71.4 \\
 & MMMU & \textbf{46.8} & 45.9 & 44.5 \\
 & OCRBench v2 & 55.0 & \textbf{55.7} & 54.3 \\
 & CharXiv & 37.55 & \textbf{41.55} & 37.15 \\
 & VisualWebBench & 55.6 & \textbf{68.4} & 59.4 \\
 & CVBench & 57.0 & \textbf{59.7} & 58.8 \\
\midrule
\multirow{6}{*}{\textit{Video}} & EgoSchema & 29.6 & \textbf{37.6} & 25.8 \\
 & LongVideoBench & 38.5 & \textbf{41.8} & 38.9 \\
 & MLVU & 44.14 & \textbf{46.94} & 43.68 \\
 & MMVU & 45.8 & \textbf{47.5} & 45.8 \\
 & VideoMME & 41.11 & \textbf{45.48} & 39.65 \\
 & TemporalBench & 49.4 & \textbf{53.4} & 50.6 \\
\midrule
\multirow{4}{*}{\textit{PPL $\downarrow$}} & TAU2 & 1.155 & \textbf{1.148} & 1.150 \\
 & AgentCompany & 1.248 & 1.249 & \textbf{1.247} \\
 & HLE & \textbf{1.275} & 1.278 & \textbf{1.275} \\
 & SWE & \textbf{1.216} & 1.218 & \textbf{1.216} \\
\bottomrule
\end{tabular}
\end{adjustbox}
}
\end{table}

To evaluate whether \method{} remains effective after long-context scaling, we conduct an additional extension experiment on the \method{}-CPT model. Starting from the sparse continued-pretraining checkpoint, we run approximately 140B tokens of long-context training and then evaluate on HELMET and RULER. The results are reported in \Tabref{tab:lctx-main}. After the extension stage, \method{}-CPT remains close to the Full-Attention baseline. Since each query and GQA group still attends to only \(k B_k=16\times128=2{,}048\) key-value tokens, these results indicate that \method{} can preserve long-context capability under a highly tight attention budget.

Additional ablations supporting these design choices are provided in the appendix. In particular, \Secref{app:prelim} studies the training recipe for the Index Branch, including gradient sources, KL-gradient detachment, warmup, and the comparison with a sliding-window sparse baseline. \Secref{app:additional-ablation} further examines architectural choices such as block size, forced sink, local selection, and the Index Branch value head. These ablations provide the empirical basis for the final \method{} design used in the main experiments.

\begin{table}[H]
\centering
\caption{Long-context extension results for \method{}-CPT on HELMET and RULER. \(\Delta\) reports the difference between \method{}-CPT and the Full-Attention baseline. The "Overall" score is averaged across the fine-grained subtasks. Higher is better for all metrics.}
\label{tab:lctx-main}
{\small
\renewcommand{\arraystretch}{0.92}
\begin{adjustbox}{max width=0.98\linewidth}
\begin{tabular}{llrrr}
\toprule
Benchmark & Subset & Full & \method{}-CPT & $\Delta$ \\
\midrule
\multirow{3}{*}{\textit{HELMET-128K}} & Overall & \textbf{46.53} & 45.93 & -0.60 \\
 & ICL & 70.40 & \textbf{72.80} & +2.40 \\
 & Rerank/RAG & \textbf{34.60} & 32.50 & -2.10 \\
\midrule
\multirow{5}{*}{\textit{RULER-128K}} & Overall & 72.00 & \textbf{72.12} & +0.12 \\
 & CWE/FWE & \textbf{46.35} & 45.00 & -1.35 \\
 & MK/MQ/MV & 96.63 & \textbf{98.87} & +2.24 \\
 & QA1/QA2 & \textbf{47.80} & 46.80 & -1.00 \\
 & VT & \textbf{97.80} & 96.80 & -1.00 \\
\bottomrule
\end{tabular}
\end{adjustbox}
}
\end{table}

\subsection{Efficiency}
\label{sec:experiment-efficiency}

We instantiate the complexity analysis in \Secref{sec:msa-flops} on our experimental model configuration and report both theoretical attention-FLOPs reduction and measured runtime speedup. Dense GQA and \method{} use the same query head count, key-value head count, head dimension, and context length; the only difference is that dense GQA attends to the full context, whereas \method{} performs index selection followed by sparse attention over a fixed KV budget. In our setting, \method{} uses $B_k=128$ and $k=16$, corresponding to a selected budget of $k B_k=2{,}048$ tokens per query.

\begin{figure}[H]
\centering
\includegraphics[width=0.98\linewidth]{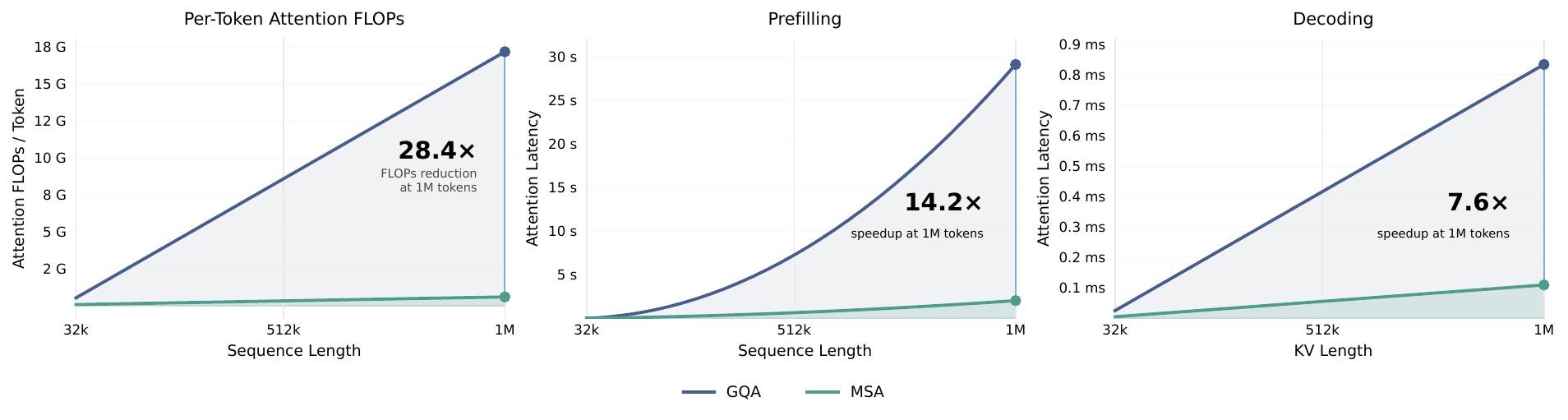}
\caption{Efficiency comparison between GQA and \method{} under the shared experimental model configuration. The left subfigure reports the theoretical per-token attention-FLOPs. The middle and right subfigures report the measured implementation speedups for prefill and decode, respectively. All tests are conducted with 64 query heads, 4 key-value heads, and a head dimension of 128. \method{} uses $B_k=128$ and $k=16$, corresponding to a selected budget of $2{,}048$ tokens per query.}
\label{fig:efficiency}
\end{figure}

As shown in \Figref{fig:efficiency}, \method{} reduces per-token attention FLOPs substantially relative to GQA in our setting, with the reduction increasing at longer contexts. At $1\mathrm{M}$ tokens, the FLOPs reduction reaches $28.4\times$ under the same head configuration.
The measured runtime speedup follows the same scaling trend but is not expected to match the FLOPs reduction exactly. Sparse attention introduces index construction, top-$k$ selection, reverse-index materialization, query gathering, and load-balancing overheads, and its memory access pattern is less regular than dense attention. Therefore, the runtime speedup is smaller than the theoretical FLOPs reduction, but it increases with context length as the dense baseline continues to scale with the full sequence while \method{} keeps the main attention budget fixed.

%% file: sections/related_works.tex
\section{Related Works}
\label{sec:related-works}

Long-context efficiency has motivated a large body of work on efficient attention, which can be broadly grouped into two directions: replacing dense softmax attention with cheaper linear or recurrent alternatives, and retaining softmax attention while restricting its receptive field.
Linear attention~\citep{katharopoulos2020linear,choromanski2020performer} replaces the softmax kernel with a linear-complexity surrogate, while state-space models such as Mamba~\citep{gu2023mamba} replace attention with a selective recurrence over hidden states.
Hybrid stacks~\citep{minimax01,minimaxm1} interleave linear blocks with full-attention blocks, reducing the number of quadratic layers while preserving part of the exact-softmax capacity.
Fixed-pattern attention keeps softmax attention but imposes a predefined support, including local windows, global tokens~\citep{beltagy2020longformer,zaheer2020bigbird}, and attention sinks with a sliding window~\citep{xiao2023streaming}.
These approaches reduce long-context cost either by replacing dense attention in part or in full, or by using a content-agnostic attention pattern.

Beyond fixed sparse patterns, adaptive sparse attention makes the attended support depend on the input. 
Existing methods differ mainly in when this support is constructed and whether the selector is trained as part of the model. 
Inference-time sparsification operates on a pretrained Full-Attention backbone and constructs sparse supports only during serving. H2O~\citep{zhang2023h2o} and SnapKV~\citep{li2024snapkv} prune the KV cache during decoding using accumulated attention statistics, Quest~\citep{tang2024quest} performs page-level importance estimation per query, MInference~\citep{jiang2024minference} and FlexPrefill~\citep{lai2025flexprefill} dispatch per-head sparse kernels at prefill, and InfLLM~\citep{xiao2024infllm} maintains attention sinks, a local context window, and retrievable chunks. These methods inherit the training cost of Full Attention and leave at least one inference phase near Full-Attention speed. Natively trained sparse-attention designs train the indexer during pretraining and form the closest prior work to \method{}. NSA~\citep{yuan2025nsa} targets MQA/MHA backbones with three parallel branches: compressed attention for coarse global context, selected attention over fine-grained blocks, and a sliding window for local context. InfLLM-V2~\citep{zhao2025infllmv2} achieves zero-shot dense-to-sparse switching by unifying parameter-free block selection with a local sliding window. MoBA~\citep{lu2025mobamixtureblockattention} also operates on GQA but uses very large KV blocks scored by block-averaged keys, and trains its indexer only through the language-modeling gradient. DSA~\citep{deepseekai2025deepseekv32pushingfrontieropen} sits on top of MLA in its MQA mode: a multi-head ReLU-based lightning indexer scores tokens individually, all query heads share a single Top-$k$ index, and selection is token-level.
\method{} differs from this neighborhood along two axes that are taken up together: per-GQA-group Top-$k$ sharing combined with block-level selection, which gives multi-group block-granular retrieval while keeping KV reads contiguous.

Efficient kernels are essential for sparse attention to translate theoretical FLOP reduction into wall-clock speedups. FlashAttention~\citep{dao2022flashattention} and FlashAttention-2~\citep{dao2023flashattention2} introduced IO-aware tiled softmax attention, and FlashDecoding~\citep{dao2023flashdecoding} extended this to memory-bound decoding. Open-source block-sparse kernels such as Flash-Sparse-Attention~\citep{yan2025fsa} and FlashMoBA~\citep{xiao2025flashmoba} have made block-sparse variants of this recurrence available. \method{}'s kernel, described in \cref{sec:infrastructure}, reuses the FlashAttention algorithmic skeleton with a loop ordering tuned to the GQA-native, block-granular access pattern \method{} produces.

%% file: sections/conclusion.tex
\section{Conclusion}
\label{sec:conclusion}

We introduced \method{}, a sparse-attention mechanism co-designed with Grouped-Query Attention. The architecture attaches a lightweight Index Branch to a standard GQA layer: each GQA group independently selects a small set of key-value blocks through a block-level dot-product indexer, and the Main Branch performs softmax attention restricted to the selected blocks. The Index Branch is a pure selector and is trained by a KL alignment loss against the Main Branch under a two-stage warmup schedule and a stop-gradient on the index input that confines the auxiliary loss to the index projections. At the 109B-MoE scale, \method{} preserves the capability of a GQA Full-Attention baseline across most pretraining and agentic benchmarks while reducing per-token attention compute by $28.4\times$ at $1\mathrm{M}$ context, the regime in which long-context inference becomes the binding deployment constraint.

\textbf{Outlook.}
\method{}'s core decisions---per-GQA-group independent selection, block-level granularity, and an indexer trained with a KL alignment objective---compose with the GQA backbone shared by most current open-source frontier models, so the recipe should transfer with little modification. Two directions are natural next steps: closing the residual long-context retrieval gap, either through longer sparse training, a larger selection budget at inference time, or a richer indexer scoring function; and extending the same selector-only design to settings beyond pretraining, including reinforcement-learning post-training and agentic deployment, where long-context cost is the dominant operational constraint.

%% file: sections/appendix.tex
\appendix

\section{Visualization}

\begin{figure}[!htb]
\centering
\begin{subfigure}[t]{0.48\linewidth}
  \centering
  \includegraphics[width=\linewidth]{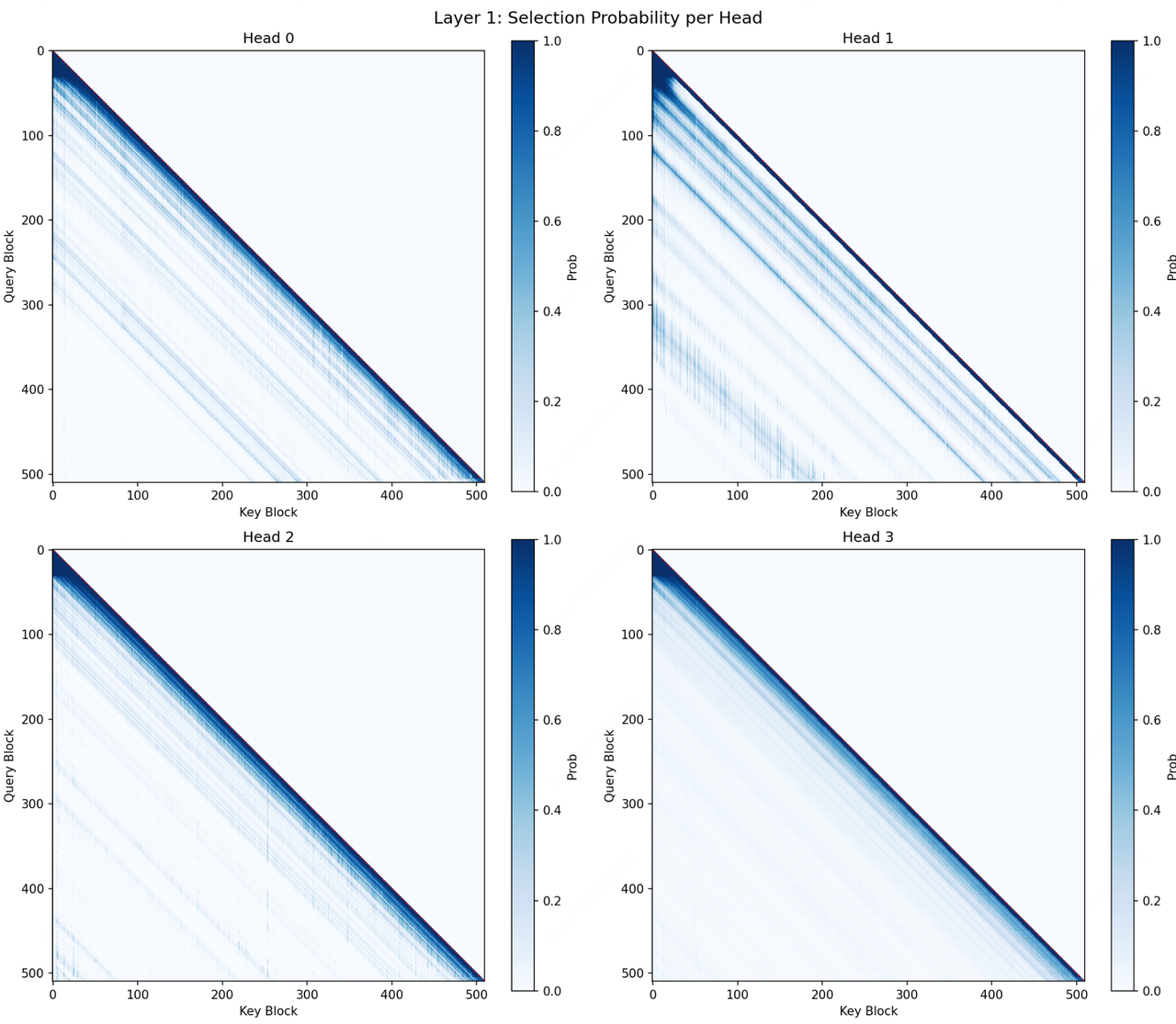}
  \caption{Layer 1, four GQA groups. Each group produces a different long-range selection pattern alongside the shared local diagonal and sink column.}
\end{subfigure}\hfill
\begin{subfigure}[t]{0.48\linewidth}
  \centering
  \includegraphics[width=\linewidth]{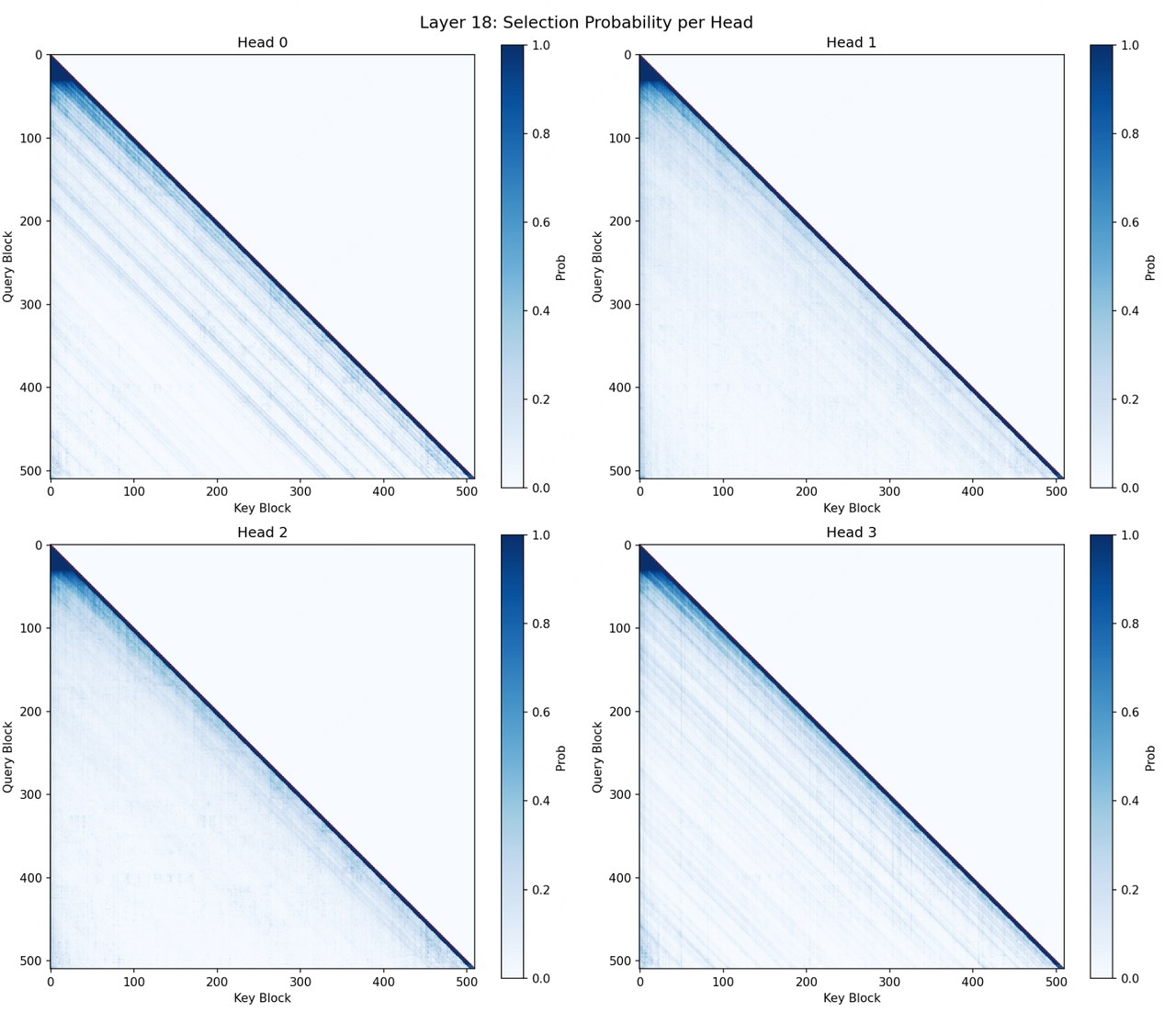}
  \caption{Layer 18, four GQA groups. Long-range selection sharpens into a few stripes per group; the four groups pick visibly different stripes.}
\end{subfigure}
\caption{Per-head Index Branch selection probability across query-key pairs. Each panel shows four heads from one layer, corresponding to four different GQA groups. All groups consistently select the local diagonal and the sink column (leftmost), while different groups trace different long-range stripes, revealing group-specific sparse selection patterns.}
\label{fig:vis-selection}
\end{figure}

To better understand what the learned indexer selects, we visualize the per-head Index Branch selection probability over all query-block and key-block pairs in \Figref{fig:vis-selection}. We show four heads from an early layer (Layer~1) and a later layer (Layer~18), corresponding to four different GQA groups. Across layers, the learned sparse pattern recovers the main structures expected from dense attention: all heads place high probability on the local diagonal, consistently select the sink column, and reserve the remaining budget for a small number of long-range relative positions. At the same time, the non-local selections are not identical across GQA groups. Different groups attend to different long-range stripes while sharing the common local and sink patterns, suggesting that the learned indexer captures group-specific sparse attention patterns rather than collapsing to a single global selection pattern.

\begin{figure}[!htb]
\centering
\includegraphics[width=0.92\linewidth]{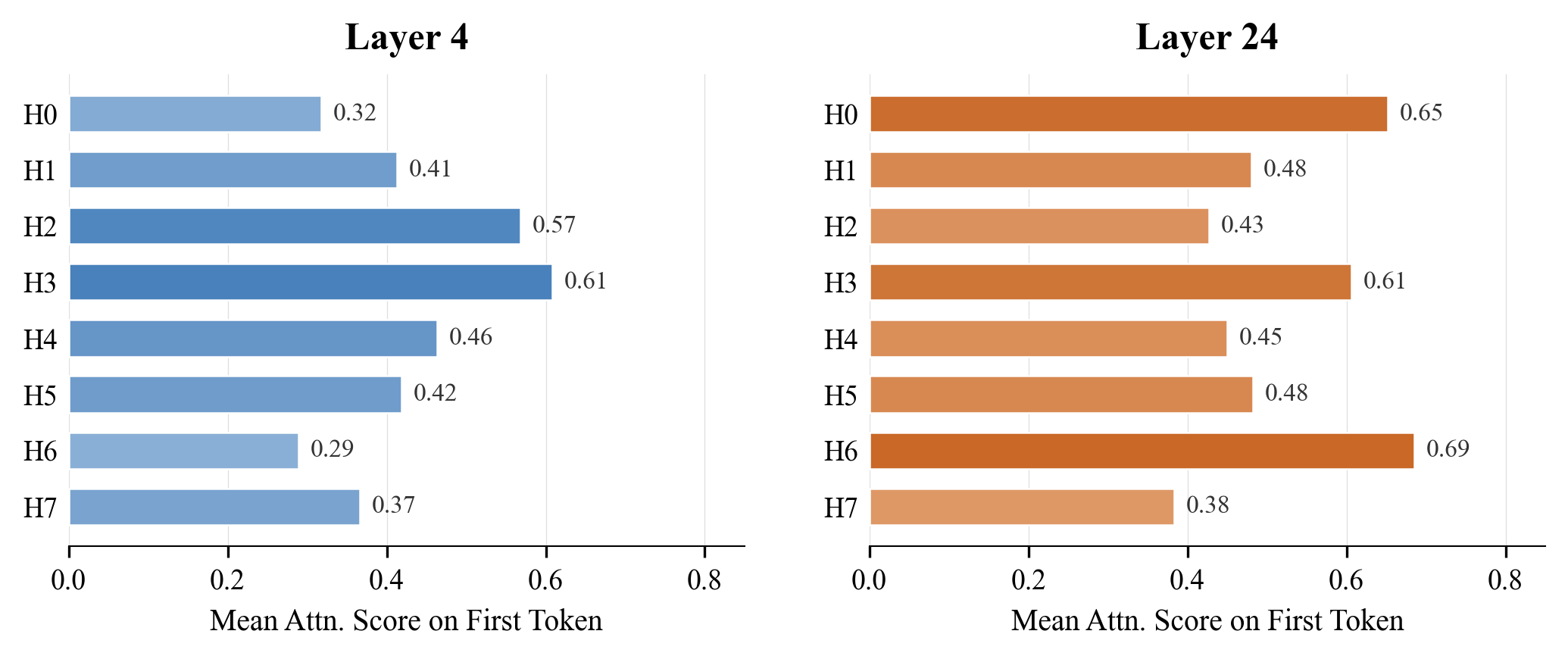}
\caption{Mean attention score on the first token for each attention head in Layer~4 and Layer~24. All heads allocate a significant fraction of attention to the first token, confirming a pervasive attention sink effect across heads and layers.}
\label{fig:attention-sink}
\end{figure}

We further examine the attention sink phenomenon in \method{} models. Even without explicitly forcing the indexer to select the first key-value block, we observe that the learned Index Branch naturally assigns high selection probability to the initial block across all layers and heads. \Figref{fig:attention-sink} shows results for two representative layers (Layer~4 and Layer~24), each with eight sampled heads. Across both layers, every head directs a substantial fraction of its attention mass to the first token. This confirms that attention focal points naturally emerge and are universally present across different heads and layers, even in our sparse attention mechanism.

\section{Preliminary Experiments}
\label{app:prelim}

This section presents small-scale ablation studies on a pilot model. Our goal is to identify the training-design choices that are essential for stable optimization and strong downstream performance. These results serve as the empirical basis for the final recipe described in \Secref{sec:method}.

\subsection{Setup}
\label{app:prelim-setup}

All ablations in this section use a 10B-parameter pilot Transformer with the same architecture family as the main paper \method{} model but with 16 layers. The model uses a 200K-token vocabulary and hidden size $d_{\rm model}=2048$. Each attention module uses GQA with 32 query heads, 4 KV heads, head dimension 128, and RoPE dimension 64. The MoE contains 64 experts with top-4 expert routing and expert inner dimension 1536. The model has 10.53B total parameters and 1.47B active parameters per token. The optimizer, learning-rate schedule, and tokenizer match the full-scale configuration. Each run is trained on a subset of the same pretraining corpus used at full scale.

\subsection{Gradient Sources for the Index Branch}
\label{app:prelim-grad-source}

A central challenge in training the Index Branch is that the top-$k$ selection in \Eqref{eq:msa-topk} is non-differentiable. Under the plain sparse-attention forward pass, the selected block indices are used only as a discrete routing decision. Consequently, the index projections $\mW^{\rm idx}_q$ and $\mW^{\rm idx}_k$ receive no useful gradient from the language-modelling objective, and the indexer cannot learn which blocks should be selected.
There are several possible ways to introduce a training signal for the indexer. We investigate two mechanisms that preserve the sparse-attention structure while providing gradients to the Index Branch.

\textbf{Index Branch output.} The first mechanism lets the Index Branch contribute an additional attention output. Specifically, we attach a value projection to the Index Branch and compute $\mO^{\rm idx}=\mathrm{Attn}(\mQ^{\rm idx}, \mK^{\rm idx}, \mV^{\rm idx})\in \mathbb{R}^{N \times H_q \times d_h}$. This output is added to the layer output through a separate output projection, $\mO'=\mW_o \mO+\mW^{\rm idx}_o \mO^{\rm idx}$. 
This design trains the Index Branch through its contribution to next-token prediction.

\textbf{KL loss.} The second mechanism directly supervises the Index Branch by matching its selection distribution to the Main Branch on the selected support. We use the auxiliary loss $\Ls_{\rm KL}$ defined in \Eqref{eq:msa-kl}. This loss acts on $\mW^{\rm idx}_q$ and $\mW^{\rm idx}_k$, and provides an explicit training signal for the index selection. 

To separate the effects of these two gradient sources, we train the model from scratch in three configurations, using sparse attention from the first step:
\begin{itemize}
    \item \textbf{LM Loss only}: the Index Branch output is added to the layer output, and the model is trained only with the language-modelling loss,
    \begin{equation}
    \mO' = \mW_o \mO + \mW^{\rm idx}_o \mO_{\rm idx},
    \qquad
    \Ls = \Ls_{\rm LM}.
    \end{equation}

    \item \textbf{KL Loss only}: the Index Branch output is discarded, and the indexer is trained only through the auxiliary KL loss,
    \begin{equation}
    \mO' = \mW_o \mO,
    \qquad
    \Ls =\Ls_{\rm LM}+\lambda \sum_{\rm layers} \Ls_{\rm KL}.
    \end{equation}

    \item \textbf{LM Loss\,+\,KL Loss}: both mechanisms are enabled,
    \begin{equation}
    \mO' = \mW_o \mO + \mW^{\rm idx}_o \mO_{\rm idx},
    \qquad
    \Ls =\Ls_{\rm LM}+\lambda \sum_{\rm layers} \Ls_{\rm KL}.
    \end{equation}
\end{itemize}

\begin{figure}[!htb]
\centering
\includegraphics[width=0.85\linewidth]{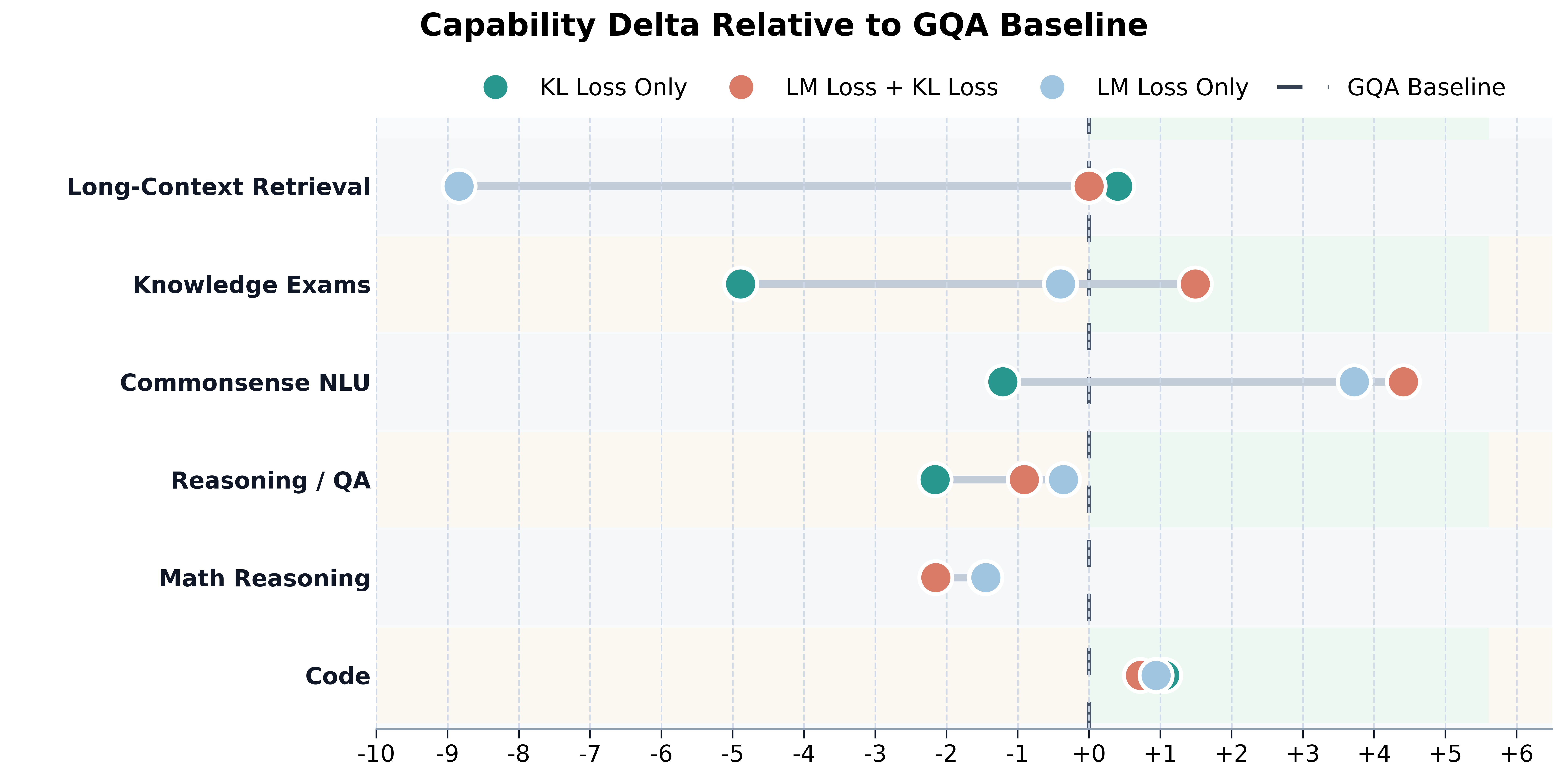}
\caption{Evaluation-score deltas relative to the Full-Attention
baseline for three indexer training signals in the pilot setting.
Positive values indicate improvements over the baseline, and negative
values indicate degradations.}
\label{fig:prelim-grad-source}
\end{figure}

\Figref{fig:prelim-grad-source} reports the per-benchmark delta of each configuration against the Full-Attention GQA baseline trained on the same data. The two single-signal configurations show complementary weaknesses. \textbf{LM Loss only} preserves short-context ability but performs poorly on long-context retrieval: without an objective on the top-$k$ selection itself, the indexer receives little direct pressure to select relevant blocks. \textbf{KL Loss only} improves retrieval but reduces short-context ability: removing $\mO_{\rm idx}$ from the layer output reduces the attention capacity available to the language model. \textbf{LM Loss\,+\,KL Loss} gives the best balance across the two axes and is the configuration we use for the remaining ablations in this section.

Based on these results, we use the \textbf{LM Loss\,+\,KL Loss} configuration for the remaining ablations in this section. We later show in \Secref{app:from-scratch-no-idx-value} that, once the indexer warmup introduced in \Secref{app:prelim-warmup} is used in the full-scale setting, the Index Branch output is no longer necessary. The final recipe, therefore, keeps the KL supervision but removes the Index Branch value head and its additive output path.

\subsection{Confining the KL Gradient to the Index Branch}
\label{app:prelim-detach}

The auxiliary KL loss is intended to train the Index Branch to match the Main Branch selection distribution. Under the default autograd graph, the KL gradient flows through the Index Branch query and key projections back into the hidden state, and then into the backbone through the residual stream. In this case, the KL loss becomes an additional objective on the backbone, rather than a local supervision signal for the indexer.

\begin{figure}[!htb]
\centering
\includegraphics[width=0.85\linewidth]{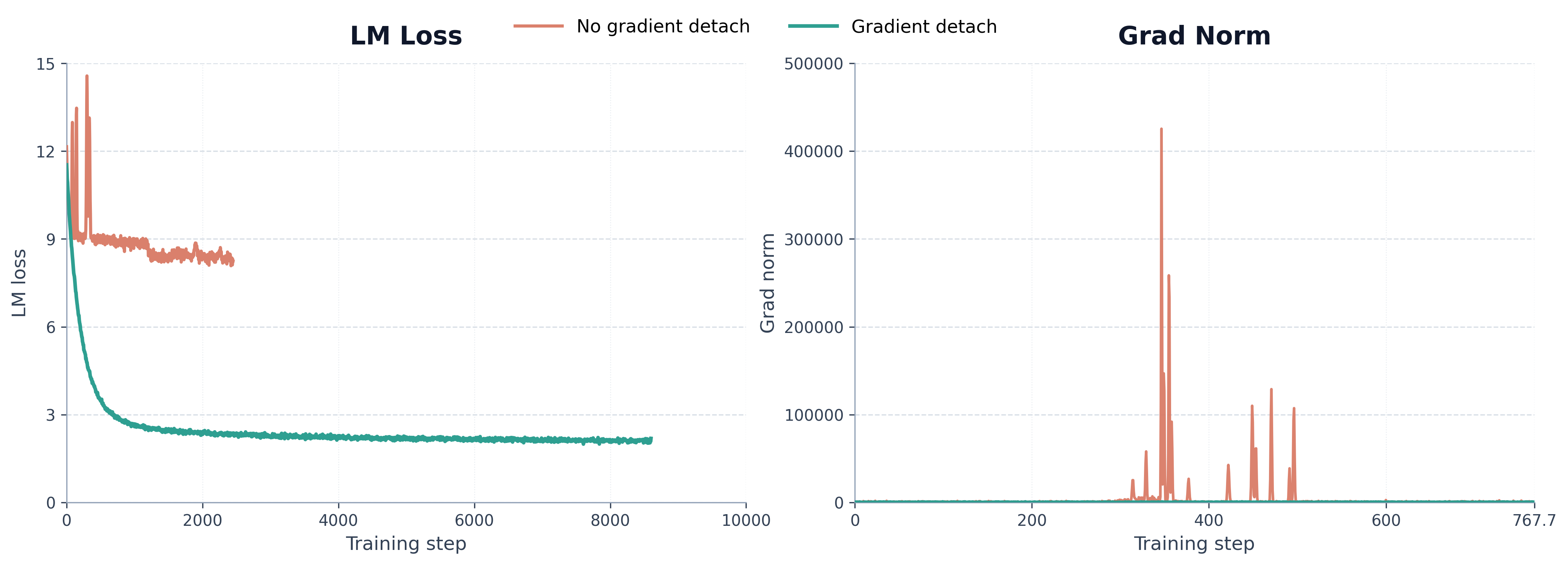}
\caption{
Training LM loss and gradient norm with and without detaching the KL gradient from the backbone. Detaching confines the auxiliary loss to the Index Branch and avoids the gradient spikes observed without detach.
}
\label{fig:prelim-detach}
\end{figure}

\begin{figure}[!htb]
\centering
\includegraphics[width=0.8\linewidth]{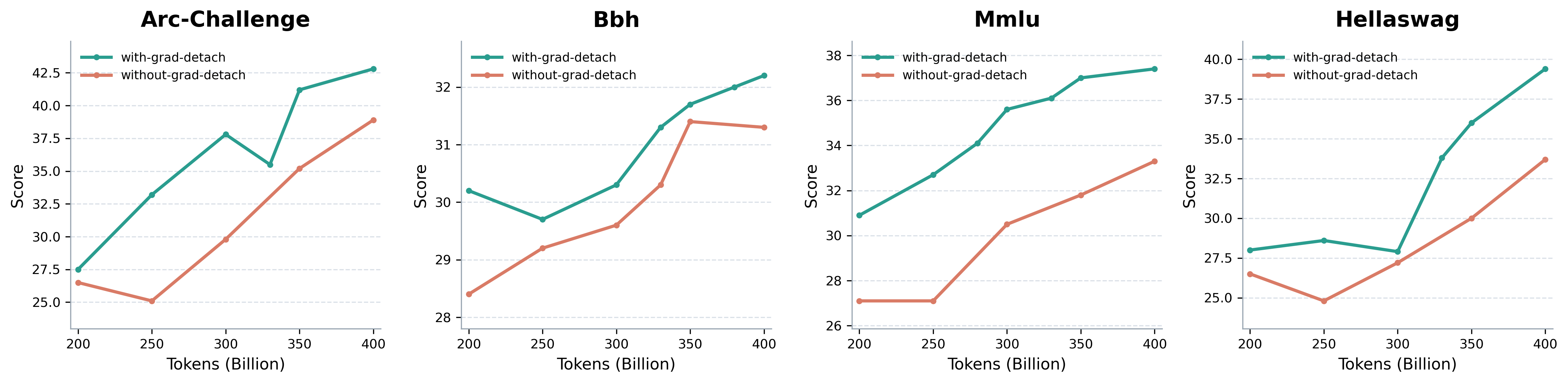}
\caption{
General benchmark scores with and without detaching the KL gradient from the backbone. Detaching the auxiliary loss reduces the general ability degeneration observed when the KL gradient updates the backbone.
}
\label{fig:prelim-detach-bench}
\end{figure}

We observe two failure modes from this gradient routing. With larger KL coefficients, occasional KL-gradient spikes propagate into the backbone, causing gradient-norm spikes and LM-loss divergence within a few hundred steps (\Figref{fig:prelim-detach}). Even at stable coefficients, standard short-context benchmarks gradually regress during training (\Figref{fig:prelim-detach-bench}). We attribute this regression to a self-distillation effect: the backbone can lower the KL loss by simplifying the Main Branch attention distribution, rather than by improving the Index Branch.

We address both failure modes by stopping the KL gradient at the Index Branch input (\Secref{sec:method-training}). Thus, each layer's KL loss becomes a local supervision signal for its own indexer. With this detach, the LM loss and gradient norm remain stable under the same KL coefficients that cause divergence without detach (\Figref{fig:prelim-detach}), and the short-context regression is removed (\Figref{fig:prelim-detach-bench}). We use this detach in all subsequent runs.

\subsection{Indexer Warmup}
\label{app:prelim-warmup}

\begin{figure}[!htb]
\centering
\includegraphics[width=0.85\linewidth]{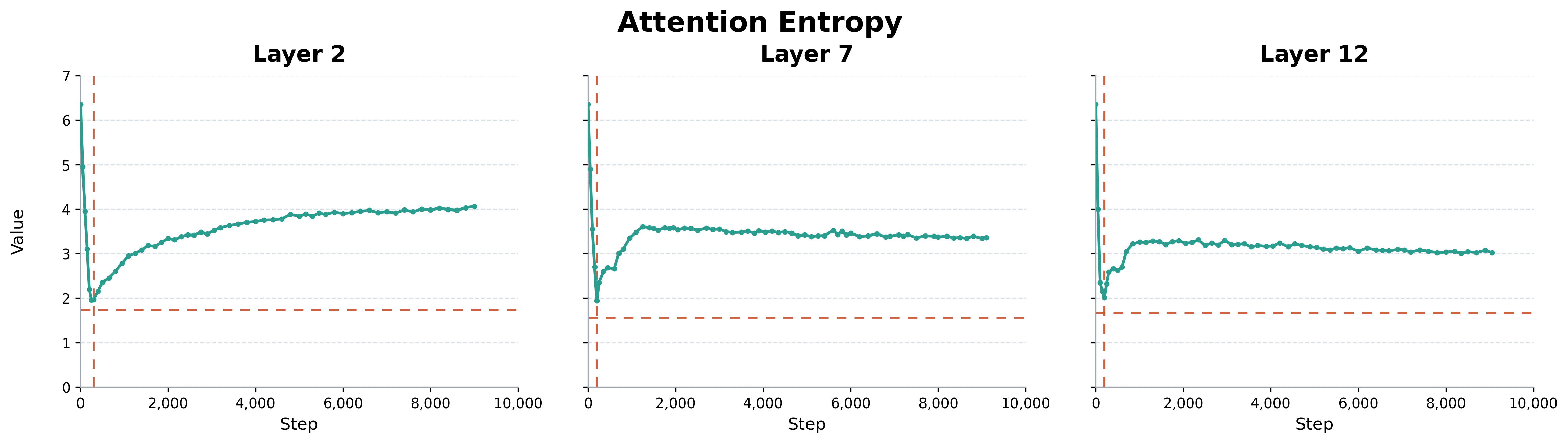}
\caption{Per-layer entropy of the Main Branch attention distribution
during early sparse training. Entropy drops rapidly in the first few
hundred steps before partially recovering and stabilizing, motivating a
brief full-attention warmup for the indexer.}
\label{fig:prelim-warmup-entropy}
\end{figure}

We observe that the Main Branch attention distribution changes rapidly during the earliest stage of training. As shown in \Figref{fig:prelim-warmup-entropy}, the attention entropy quickly drops from an initially smooth distribution to a much sharper one, before entering a slower phase of representation learning. This makes sparse selection fragile at initialization. If top-$k$ selection is enabled from step zero, the Index Branch must track a rapidly moving target while its own selections are still nearly random. Poor early selections then route the Main Branch to uninformative tokens, which weakens both backbone learning and the KL supervision received by the indexer.

We address this issue with a short indexer warmup. During warmup, the Main Branch uses full attention, while the Index Branch is trained by the KL loss against the full-sequence Main Branch distribution. This allows the backbone to pass through the early sharpening phase without sparse routing errors, and gives the indexer a meaningful initialization before it controls token selection. After $T_{\rm warm}$ steps, we enable top-$k$ sparse selection and continue training with the KL loss restricted to the selected support.

\begin{figure}[!htb]
\centering
\includegraphics[width=0.98\linewidth]{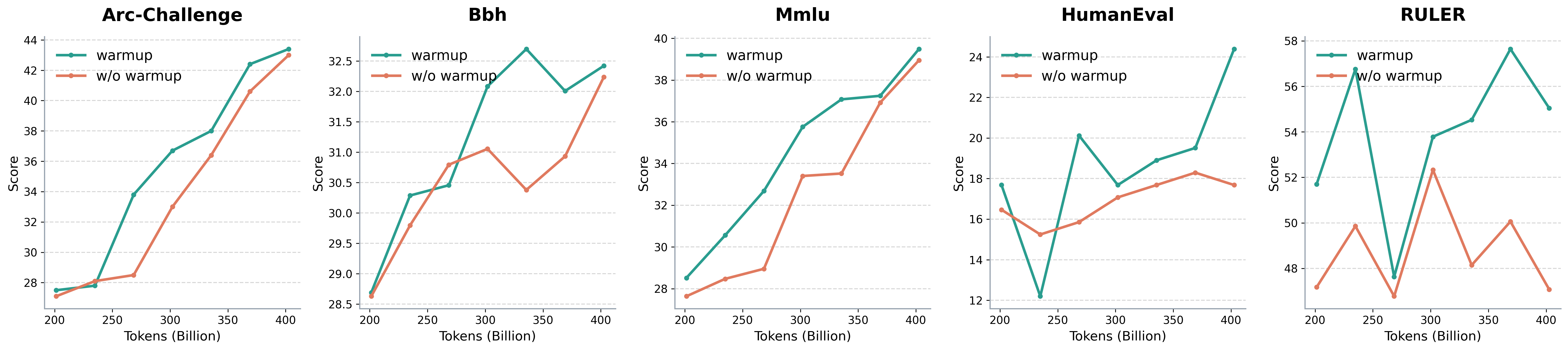}
\caption{Evaluation results of \method{} with and without index warmup. Within the reported training range, index warmup improved scores on general tasks and long-context retrieval.}
\label{fig:prelim-warmup-curves}
\end{figure}

\Figref{fig:prelim-warmup-curves} compares pretraining runs with and without this warmup. The warmed-up run achieves better short-context performance and stronger long-context retrieval. These results indicate that a short full-attention warmup provides a better initialization for sparse training. We therefore also adopt this warmup when converting Full-Attention checkpoints to sparse attention through continued pretraining.

\subsection{Learnable Attention Sink}
\label{app:prelim-learnable-sink}

The visualization in \Figref{fig:attention-sink} shows that the first token often acts as an attention sink: many heads assign a non-trivial amount of attention mass to the sequence prefix, even when the sparse selector is not explicitly forced to include it. This raises the question of whether this sink behavior should be represented by an explicit learnable mechanism, rather than being absorbed by the first real token in the sequence.
We therefore tested a GPT-OSS-style learnable attention sink. Concretely, each attention head is given an additional learnable sink logit, which competes with normal key positions in the attention softmax.

\begin{figure}[!htb]
\centering
\includegraphics[width=0.92\linewidth]{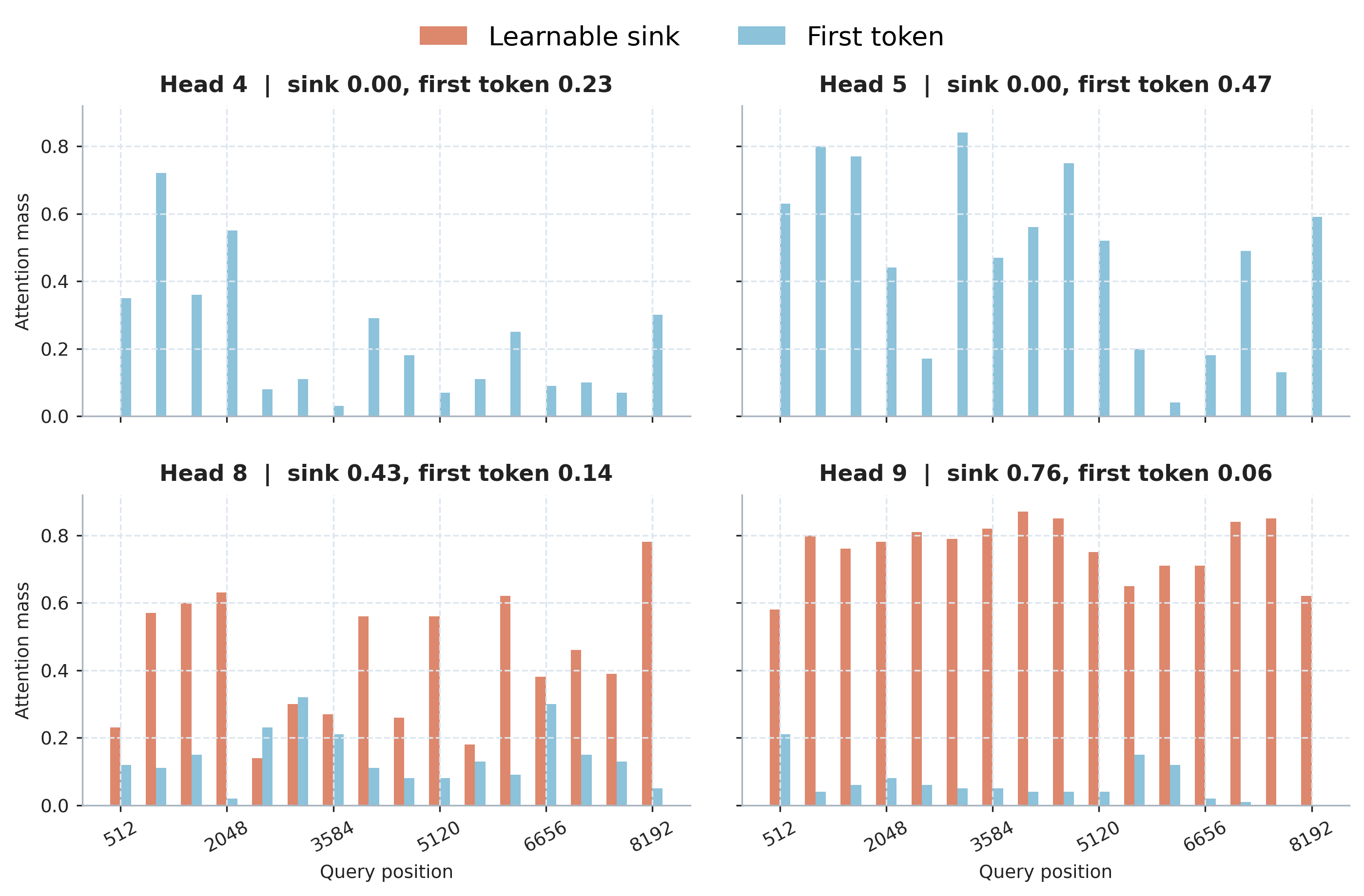}
\caption{Attention received by the learnable sink and the first token after introducing a GPT-OSS-style sink parameter. In some heads, the learnable sink absorbs most of the sink-like attention; in others, the first token remains the dominant sink, indicating that the explicit sink does not fully eliminate first-token sink behavior.}
\label{fig:prelim-learnable-sink-vis}
\end{figure}

\Figref{fig:prelim-learnable-sink-vis} visualizes the resulting attention patterns. The learnable sink absorbs substantial attention mass in some heads, but it does not completely remove the original first-token sink. In several heads, especially those where the learned sink receives little mass, the first token still receives substantial attention and continues to behave as an implicit sink.

\begin{figure}[!htb]
\centering
\includegraphics[width=0.85\linewidth]{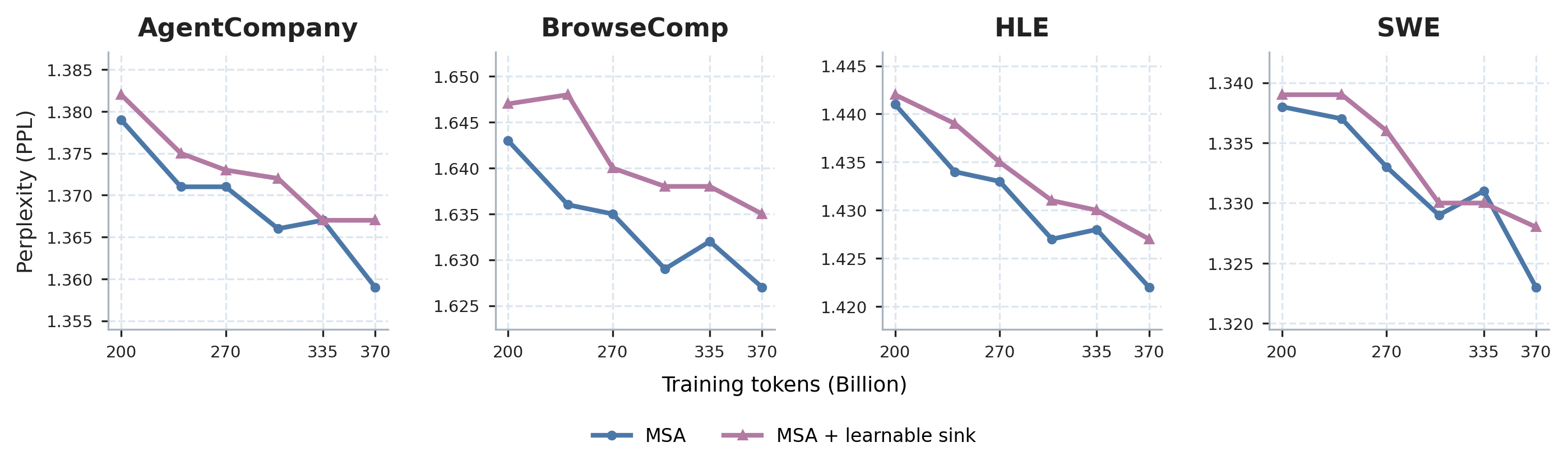}
\caption{Perplexity comparison with and without the learnable attention sink on downstream agent-oriented evaluations. Lower perplexity is better. Adding the learnable sink does not provide a consistent advantage over the default \method{} design.}
\label{fig:prelim-learnable-sink-results}
\end{figure}

We also compare downstream perplexity with and without the learnable sink in \Figref{fig:prelim-learnable-sink-results}. The learnable-sink variant does not yield a clear or consistent improvement over the default design. Given its additional parameters, implementation complexity, and the fact that it does not fully suppress first-token sink behavior, we do not include the learnable attention sink in the final recipe.

\subsection{Dynamic Sparse Selection vs.\ Sliding Window}
\label{app:prelim-swa}

To assess the value of dynamic selection, we compare \method{} with a FLOP-matched sliding-window baseline. This baseline removes the Index Branch and uses a fixed sparse pattern: each query attends to the first key block and to a local window with the same token budget ending at the query. Therefore, the two methods have the same selection budget and differ only in whether the selected tokens are fixed by position or chosen dynamically.

\begin{figure}[!htb]
\centering
\includegraphics[width=0.9\linewidth]{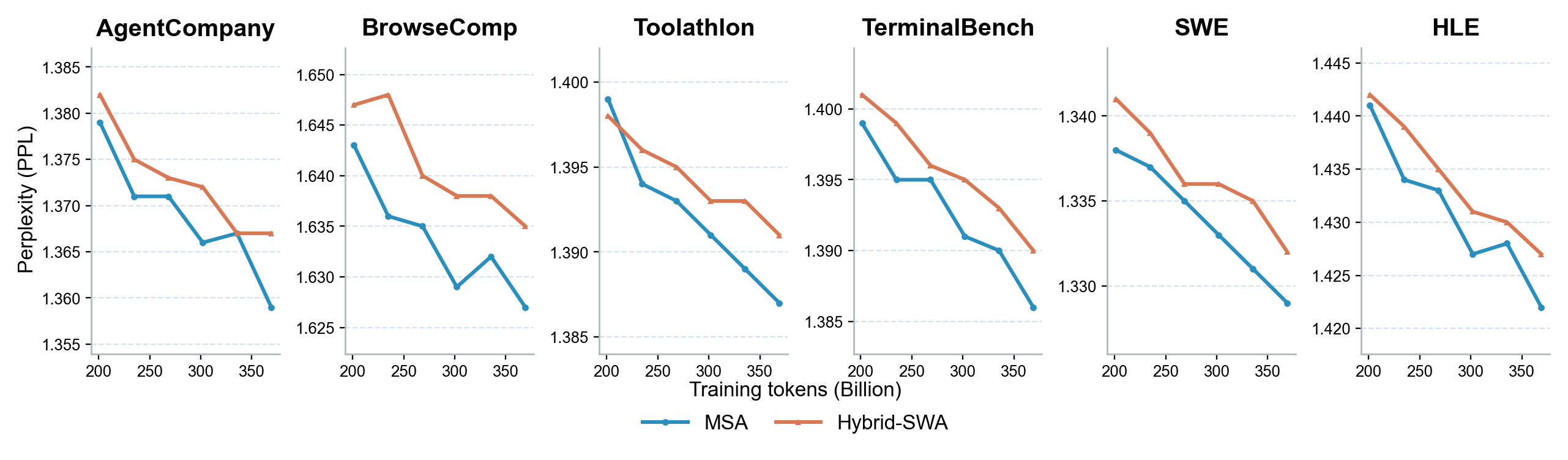}
\caption{Perplexity comparison between \method{} and a FLOP-matched
sliding-window baseline on downstream agent-oriented evaluations. Lower
Perplexity indicates better modeling performance under the same sparse
selection budget.}
\label{fig:prelim-swa-ppl}
\end{figure}

Figure~\ref{fig:prelim-swa-ppl} reports perplexity on downstream agent tasks. Under the same sparse selection budget, the sliding-window model has higher perplexity than \method{} across the training trajectory. Although both models benefit from additional training tokens, the fixed local-window pattern does not match the perplexity of dynamic sparse selection. This suggests that, for these agent tasks, a position-fixed sparse pattern is less suitable than content-dependent token selection.

\section{Additional Ablation Study}
\label{app:additional-ablation}

\subsection{Block Size}
\label{app:prelim-block-size}

The sparse attention calculation in \method{}'s Main Branch processes key-value pairs in units of consecutive $B_k$ token blocks, which affects both model performance and efficiency. Larger blocks can improve kernel efficiency but may reduce retrieval quality because of coarser selection granularity. By adjusting $B_k$ while keeping the total number of selected tokens constant, we investigate this trade-off. Compared to the main experiment, these runs use fewer training iterations and a subset of the evaluation suite.

As shown in Table~\ref{tab:block-ablation}, varying the block size has a limited impact on model quality in this setting. The PPL results are nearly unchanged across different $B_k$ values, and the RULER scores show no clear degradation when increasing the block size from 32 to 64 or 128. This suggests that \method{} can use larger key-value blocks to improve kernel efficiency with limited quality loss in these ablations.

\begin{table}[!htb]
\centering
\caption{Perplexity and long-context retrieval scores for different
key-value block sizes. Lower is better for perplexity, and higher is
better for RULER scores.}
\label{tab:block-ablation}
\small
\begin{tabular}{l c c c}
\toprule
Benchmark & Block 32 & Block 64 & Block 128 \\
\midrule
\multicolumn{4}{l}{\textit{PPL $\downarrow$}} \\
TAU2                             & 1.176 & 1.176 & 1.176 \\
AgentCompany                     & 1.266 & 1.276 & 1.266 \\
HLE                              & 1.299 & 1.299 & 1.300 \\
SWE                              & 1.233 & 1.233 & 1.233 \\
\midrule
\multicolumn{4}{l}{\textit{Long-context retrieval}} \\
RULER-8K                          & 72.5 & 72.8 & 73.8 \\
RULER-32K                          & 66.1 & 65.3 & 64.6 \\
\bottomrule
\end{tabular}
\end{table}

\subsection{Forced Sink \& Local Selection}
\label{app:prelim-sink-local}

In early sparse-training experiments, we explicitly forced the selector to include two types of blocks: the first block in the sequence and a fixed local window around the query position. The first block corresponds to the common attention-sink pattern, while the local window preserves nearby context that is important for short-range modeling and provides dense supervision for the indexer. This design was mainly introduced as a stabilization mechanism: before the indexer becomes reliable, forcing these blocks reduces the chance that the sparse branch misses basic context during early training.

We later found that these priors do not need to be hard-coded. When the forced selection of the first block and the fixed local window are removed, the trained model still exhibits both structures: attention concentrates on the sequence prefix when useful, and nearby tokens remain frequently selected.
As shown in Table~\ref{tab:forced-sink-local}, removing forced sink and fixed local selection has little effect on standard model quality: reasoning, code, and PPL metrics remain nearly unchanged. Long-context retrieval is also comparable.
These results indicate that the sparse model can learn sink and local-selection patterns without hard-coded selection rules. Therefore, the final recipe does not force the first block or a large local window, and only forces the special incomplete self block.

\begin{table}[!htb]
\begin{minipage}[t]{0.48\textwidth}
\centering
\caption{Ablation of forced sink and local-window selection. Higher is better unless marked $\downarrow$.}
\label{tab:forced-sink-local}
\small
\begin{tabular}{l c c}
\toprule
Benchmark & No Forced & Forced \\
\midrule
\multicolumn{3}{l}{\textit{General knowledge \& reasoning}} \\
MMLU                                & 60.5 & 60.5 \\
MMLU-Pro                            & 32.5 & 33.4 \\
BBH                                 & 58.2 & 58.2 \\
ARC Challenge                       & 78.1 & 77.9 \\
\midrule
\multicolumn{3}{l}{\textit{Math}} \\
GSM8K                               & 66.0 & 66.9 \\
MGSM                                & 35.8 & 36.3 \\
\midrule
\multicolumn{3}{l}{\textit{Code}} \\
EvalPlus                            & 54.0 & 53.6 \\
BigCodeBench                        & 35.6 & 35.7 \\
MultiPL-E MBPP P@10                 & 80.1 & 79.5 \\
\midrule
\multicolumn{3}{l}{\textit{Image}} \\
ChartQA                            & 73.5 & 73.7 \\
MMMU                               & 43.6 & 42.9 \\
\midrule
\multicolumn{3}{l}{\textit{Video}} \\
VideoMMMU                          & 32.1 & 32.0 \\
\midrule
\multicolumn{3}{l}{\textit{PPL $\downarrow$}} \\
TAU2                                & 1.175 & 1.175 \\
AgentCompany                        & 1.268 & 1.266 \\
HLE                                 & 1.301 & 1.300 \\
SWE                                 & 1.235 & 1.233 \\
\midrule
\multicolumn{3}{l}{\textit{Long-context retrieval}} \\
RULER-8K                             & 71.6 & 71.7 \\
RULER-32K                            & 61.5 & 65.8 \\
\bottomrule
\end{tabular}
\end{minipage}\hfill
\begin{minipage}[t]{0.46\textwidth}
\centering
\caption{Continued pre-training ablation of the Index Branch value head.}
\label{tab:idxval-ablation}
\small
\begin{tabular}{l c c}
\toprule
Benchmark & With-value & No-value \\
\midrule
\multicolumn{3}{l}{\textit{General knowledge \& reasoning}} \\
MMLU                                & 66.4 & 67.3 \\
MMLU-Pro                            & 39.0 & 39.1 \\
BBH                                 & 65.3 & 65.9 \\
ARC Challenge                       & 82.2 & 82.4 \\
\midrule
\multicolumn{3}{l}{\textit{Math}} \\
GSM8K                               & 77.6 & 76.4 \\
MathVista                           & 45.2  & 43.6  \\
MGSM                                & 48.4 & 47.6 \\
\midrule
\multicolumn{3}{l}{\textit{Code}} \\
HumanEval                           & 60.4 & 59.1 \\
EvalPlus                            & 57.7 & 58.7 \\
BigCodeBench                        & 46.0 & 44.0 \\
\midrule
\multicolumn{3}{l}{\textit{Image}} \\
AI2D                                & 69.3 & 70.4 \\
ChartQA                             & 75.3 & 74.9 \\
MMMU                                & 44.9 & 43.4 \\
OCRBench v2                         & 53.2 & 53.9 \\
\midrule
\multicolumn{3}{l}{\textit{Video}} \\
MLVU                                & 42.4 & 43.9 \\
MMVU                                & 44.9  & 43.7  \\
PerceptionTest                      & 45.0 & 47.3 \\
\midrule
\multicolumn{3}{l}{\textit{Long-context retrieval}} \\
RULER-8K                            & 84.1 & 83.0 \\
RULER-32K                           & 79.7 & 80.4 \\
\bottomrule
\end{tabular}
\end{minipage}
\end{table}

\subsection{Index Branch Value Head}
\label{app:from-scratch-no-idx-value}

Our preliminary experiments (\Secref{app:prelim-grad-source}) show that providing an additional attention output through the Index Branch helps the model begin sparse training from step zero. However, this index value head introduces additional computation and complexity. Since the indexer warmup in \Secref{app:prelim-warmup} already improves the initialization for sparse training, we further ablate whether the value head is still needed.

We compare the original with-value design against a no-value variant that trains the indexer only with the KL alignment signal. As shown in Table~\ref{tab:idxval-ablation}, removing the index value head does not lead to a systematic degradation across the evaluation suite. The no-value variant is slightly better on some general reasoning benchmarks, while the with-value variant retains small advantages on some math and code tasks. On multimodal benchmarks and long-context retrieval, the differences are also mixed.

Overall, the results indicate that the index value head is not critical once the Index Branch warmup is used. Its effect on downstream quality is small and benchmark-dependent, with neither variant consistently dominating the other.
This suggests that the main role of $\mO_{\rm idx}$ in the earlier recipe was to provide an additional early training signal, rather than to supply essential capacity at convergence.
The final design, therefore, drops the index value head on efficiency grounds. At inference time, the top-$k$ indexer only needs the block-wise maximum of $\mQ_{\rm idx}\mK_{\rm idx}^{\top}$, avoiding the value aggregation path and exponential calculations entirely.